\documentclass[11pt]{article}
\usepackage[a4paper,top=3cm,bottom=2cm,left=2cm,right=2cm,marginparwidth=1.75cm]{geometry}

\usepackage[numbers]{natbib}
\usepackage[colorlinks=True,citecolor=blue,urlcolor=blue,pagebackref=true,backref=true]
{hyperref}
\renewcommand*{\backrefalt}[4]{%
    \ifcase #1 \footnotesize{(Not cited.)}%
    \or        \footnotesize{(Cited on page~#2.)}%
    \else      \footnotesize{(Cited on pages~#2.)}%
    \fi}

\usepackage{rotating}
\usepackage[normalem]{ulem}
\usepackage[none]{hyphenat}
\usepackage{breakcites}
\usepackage[utf8]{inputenc} 
\usepackage[T1]{fontenc}    
\usepackage{hyperref}       
\hypersetup{
  colorlinks = true,
  urlcolor = blue, 
  linkcolor = blue,
  citecolor = blue
}
\usepackage{booktabs}       
\usepackage{amsfonts}       
\usepackage{nicefrac}       
\usepackage{microtype}      
\usepackage{xcolor}         
\usepackage{caption}
\usepackage{subcaption}
\usepackage{float}
\usepackage{amsmath,amsthm}
\usepackage{amssymb}
\usepackage{mathtools}
\usepackage{soul}
\usepackage{bm}
\usepackage{enumitem}
\usepackage{multirow}
\usepackage{wrapfig}
\usepackage{scalerel}
\allowdisplaybreaks
\usepackage{dsfont}
\usepackage[noend]{algpseudocode}
\usepackage{xcolor}
\usepackage{thmtools,thm-restate}
\usepackage{cleveref}  
\usepackage{autonum}
\usepackage[linesnumbered,ruled,algo2e]{algorithm2e}%

\newtheorem{remark}{Remark}
\newtheorem{result}{Result}

\newtheorem{definition}{Definition}

\newtheorem{assumption}{Assumption}
\newtheorem{theorem}{Theorem}
\newtheorem{corollary}{Corollary}
\newtheorem{lemma}{Lemma}
\newtheorem{proposition}{Proposition}

\newcommand{\msf}{\mathsf}
\newcommand{\rownnlink}{\hyperref[algo:row_star_NN]{\textsc{AWNN}}\xspace}
\newcommand{\rowimputelink}{\hyperref[algo:row_imputer]{\textsc{Row-imputer}}\xspace}

\newcommand{\wgtadjlink}{\hyperref[algo:wgt_adjuster]{\textsc{Weight-Adjuster}}\xspace}
\newcommand{\NR}[1][j]{\mathcal{R}\parenth{i, #1}}

\newcommand{\NC}[1][i]{\mathcal{C}\parenth{#1}}

\newcommand{\awnn}{\mathsf{AWNN}}
\newcommand{\oawnn}{\mathsf{O}-\mathsf{AWNN}}
\newcommand{\K}[1][j]{K_{i, #1}}

\newcommand{\miss}[1][i]{A_{#1, j}}

\newcommand{\distroworacle}[1][i']{\rho_{#1,i}}
\newcommand{\distrow}[1][i']{\widehat{\rho}_{i,#1}}
\newcommand{\distrowerror}[1][i']{\zeta_{#1,i}(m, \delta)}

\newcommand{\drow}[1][i]{\widehat{\rho}^2\parenth{#1,i_0}}
\newcommand{\drown}[2]{\widehat{\rho}_{#1,#2}}
\newcommand{\tdrown}[2]{\tilde{\rho}_{#1,#2}}
\newcommand{\drownmean}[1]{\Bar{\widehat{\rho}}_{i;#1}}

\newcommand{\pseqann}[1][n]{(a_i)_{i=1}^{#1}} 

\newcommand{\pseqnn}[2]{(#1_i)_{i=1}^{#2}} 

\newcommand{\wgtvector}[1][i]{w^{(#1)}}

\newcommand{\simpleth}[1][i]{\theta_{#1,j}}
\newcommand{\simplewgt}[1][i']{w_{#1}}
\newcommand{\optwgt}{w^\star}
\newcommand{\optwgtj}[1][j]{w^\star_{#1}}
\newcommand{\optlambda}{\lambda^\star}
\newcommand{\hatsigmasq}{\what\sigma^2}
\newcommand{\deno}{2C\hatsigmasq}

\newcommand{\neighbors}[1][i]{\mbf{N}_{#1}^{\mathsf{row}}}

\newcommand{\knn}{k-\mrm{NN}}

\newcommand{\sumn}[1][i]{\sum_{#1=1}^n}
\newcommand{\summn}[1][i]{\sum_{#1=1}^m}

\newcommand{\x}{x}

\newcommand{\axi}[1][i]{\x_{#1}}

\newcommand{\eps}{\epsilon}

\newcommand{\ph}{\theta}

\newcommand{\pseqxn}[1][n]{(\axi[i])_{i\geq 1}} 
\newcommand{\pseqxnn}[1][n]{(\axi[i])_{i=1}^n} 

\newcommand{\brackets}[1]{\left[ #1 \right]}

\newcommand{\parenth}[1]{\left( #1 \right)}

\newcommand{\braces}[1]{\left\{ #1 \right \}}

\newcommand{\biggbraces}[1]{\bigg\{ #1 \bigg \}}
\newcommand{\abss}[1]{\left| #1 \right |}

\newcommand{\real}{\ensuremath{\mathbb{R}}}

\def\balign#1\ealign{\begin{align}#1\end{align}}
\def\baligns#1\ealigns{\begin{align*}#1\end{align*}}
\def\balignat#1\ealign{\begin{alignat}#1\end{alignat}}
\def\balignats#1\ealigns{\begin{alignat*}#1\end{alignat*}}
\def\bitemize#1\eitemize{\begin{itemize}#1\end{itemize}}
\def\benumerate#1\eenumerate{\begin{enumerate}#1\end{enumerate}}

\newenvironment{talign*}
 {\csname align*\endcsname}
 {\endalign}
\newenvironment{talign}
 {\csname align\endcsname}
 {\endalign}

\def\balignst#1\ealignst{\begin{talign*}#1\end{talign*}}
\def\balignt#1\ealignt{\begin{talign}#1\end{talign}}

\newcommand{\qtext}[1]{\quad\text{#1}\quad}

\let\originalleft\left
\let\originalright\right
\renewcommand{\left}{\mathopen{}\mathclose\bgroup\originalleft}
\renewcommand{\right}{\aftergroup\egroup\originalright}

\def\Holder{H\"older\xspace}

\def\tinycitep*#1{{\tiny\citep*{#1}}}
\def\tinycitealt*#1{{\tiny\citealt*{#1}}}
\def\tinycite*#1{{\tiny\cite*{#1}}}
\def\smallcitep*#1{{\scriptsize\citep*{#1}}}
\def\smallcitealt*#1{{\scriptsize\citealt*{#1}}}
\def\smallcite*#1{{\scriptsize\cite*{#1}}}

\def\mbf#1{\mathbf{#1}}
\def\mbb#1{\mathbb{#1}}

\def\mrm#1{\mathrm{#1}}
\def\trm#1{\textrm{#1}}
\def\tbf#1{\textbf{#1}}

\def\<{\left\langle} 
\def\>{\right\rangle}

\def\iff{\Leftrightarrow}
\def\implies{\quad\Longrightarrow\quad}

\def\norm#1{\left\|{#1}\right\|} 
\newcommand{\twonorm}[1]{\norm{#1}_2} 
\newcommand{\infnorm}[1]{\norm{#1}_{\infty}} 

\def\what#1{\widehat{#1}}

\def\indic#1{\mbb{I}\left[{#1}\right]} 

\def\bigO#1{\mathcal{O}(#1)} 

\def\Var{\mrm{Var}} 
\def\Vararg#1{\Var\left[{#1}\right]}

\newcommand{\Gsn}{\mathcal{N}}

\newcommand{\Unif}{\textnormal{Unif}}

\newcommand{\pderiv}[2]{\frac{\partial #1}{\partial #2}} 

\newcommand{\iid}{\textrm{i.i.d.}\xspace}

\ifdefined\nonewproofenvironments\else
\ifdefined\ispres\else

\newenvironment{proof-sketch}{\noindent\textbf{Proof Sketch}
  \hspace*{1em}}{\qed\bigskip\\}
\newenvironment{proof-idea}{\noindent\textbf{Proof Idea}
  \hspace*{1em}}{\qed\bigskip\\}
\newenvironment{proof-of-lemma}[1][{}]{\noindent\textbf{Proof of Lemma {#1}}
  \hspace*{1em}}{\qed\\}
\newenvironment{proof-of-theorem}[1][{}]{\noindent\textbf{Proof of Theorem {#1}}
  \hspace*{1em}}{\qed\\}
\newenvironment{proof-attempt}{\noindent\textbf{Proof Attempt}
  \hspace*{1em}}{\qed\bigskip\\}

\newtheorem{myproposition}{Proposition}

\newtheorem{mycorollary}{Corollary}

\newtheorem{mylemma}{Lemma}

\newtheorem{mydefinition}{Definition}

\newtheorem{myremark}{Remark}

\newtheorem{myassumption}{Assumption}

 \crefname{appendix}{App.}{App.}
\crefname{equation}{}{}
\crefname{lemma}{Lem.}{Lem.}
\crefname{theorem}{Thm.}{Thm.}
\crefname{Corollary}{Cor.}{Cors.}
\crefname{algorithm}{Alg.}{Algs.}

\crefname{section}{Sec.}{Sec.}
\crefname{table}{Tab.}{Tab.}
\crefname{remark}{Rem.}{Rem.}
\crefname{definition}{Def.}{Def.}
\crefname{Proposition}{Prop.}{Prop.}
\crefname{myremark}{Rem.}{Rem.}
\crefname{mylemma}{Lem.}{Lem.}
\crefname{mydefinition}{Def.}{Defs.}
\crefname{myproposition}{Prop.}{Prop.}
\crefname{mycorollary}{Cor.}{Cors.}
\crefname{myassumption}{Assum.}{Assum.}
\crefname{figure}{Fig.}{Fig.}
\crefname{enumi}{}{}
\crefname{name}{}{} 

\title{Adaptively-weighted Nearest Neighbors for Matrix Completion}
\author{%
  Tathagata Sadhukhan\textsuperscript{1}\thanks{These authors contributed equally.}%
  \and
  Manit Paul\textsuperscript{2}\footnotemark[1]%
  \and
  Raaz Dwivedi\textsuperscript{1,3}
}
\date{%
  \textsuperscript{1}Cornell University \quad
  \textsuperscript{2} The Wharton School \quad 
  \textsuperscript{3} Cornell Tech
}

\begin{document}
\maketitle
\begin{abstract}
In this technical note, we introduce and analyze $\awnn$ -- an adaptively weighted nearest neighbor method for performing matrix completion. Nearest neighbor (NN) methods are widely used in missing data problems across multiple disciplines such as in recommender systems and for performing counterfactual inference in panel data settings. Prior works have shown that in addition to being very intuitive and easy to implement, NN methods enjoy nice theoretical guarantees. However, the performance of majority of the NN methods rely on the appropriate choice of the radii and the weights assigned to each member in the nearest neighbor set and despite several works on nearest neighbor methods in the past two decades, there does not exist a systematic approach of choosing the radii and the weights without relying on methods like cross-validation. $\awnn$ addresses this challenge by judiciously balancing the bias–variance trade-off inherent in weighted nearest-neighbor regression. We provide theoretical guarantees for the proposed method under minimal assumptions and support the theory via synthetic experiments. 
\end{abstract}

\section{Introduction}
\label{sec:intro}
Where there's a good resemblance between datapoints, there's a reason. For instance, in streaming platforms, a user's future movie ratings can be predicted by comparing the user's watching profile to the similar users, the core idea behind recommender systems~\cite{nikolakopoulos2021trust}. Additionally, in econometrics a key quantity of interest is the effect of a socio-economic policy, and often state administrations adopt it at different times, colloquially called staggered adoption~\cite{athey2022design} in causal inference. The standard way of measuring the policy effect in a state is first creating a "synthetic proxy" of that state using similar states where policy was not yet passed, then comparing outcomes (related to the policy) of the state and its proxy~\cite{abadie1, abadie2, ben2022synthetic}. These are two of many examples where similar datapoints or nearest neighbors are extensively used for inference and prediction. Nearest neighbors is a classical tool in various domains, tracing back to the $11^{th}$ century's renowned scholar Alhazen's work, Book of Optics~\cite{pelillo2014alhazen, al2015retrospect}. It employs nearest neighbors based ideas in its arguments for proposing modern (and now accepted) theory of vision which states that the lights entering the eye creates vision, countering the contemporary theory and belief (laid down by Euclid in his Optica) that light is emitted from eyes for vision. 

Despite being a millennium-old principle, nearest neighbors remains a formidable method owing to its computational scalability and flexibility in terms of minimal model assumptions for getting guarantees on its performance. Nearest neighbors is a popular choice for non-parametric regression, with applications in pattern recognition~\cite{fix1952discriminatory, cover1967nearest, cannings2020local}. After seeing huge empirical success, nearest neighbors has paved its way into modern fields like matrix completion. Matrix completion, the art of estimating the true underlying matrix from a data matrix of noisy and missing entries~\cite{recht2011simpler,Chatterjee15,chen2020noisy}, serves as a cornerstone of machine learning for addressing missing data challenges, ubiquitous in causal inference. It has been a canonical tool for recommender systems~\cite{koren2009matrix, agarwal2021causal}.

Nearest Neighbors based regression broadly is of two types, $k$-nearest neighbors ($\knn$) regression and kernel/Nadaraya Watson (NW) regression with a bandwidth $h$. Fitting the nearest neighbors regression requires prudent selection of two components: $(1)$ the number of nearest neighbors $k$ (or bandwidth $h$ for NW), and $(2)$ the weights assigned to the nearest neighbors. For proper tuning of $k$ (or $h$), prior literature still relies on procedures that are either ad-hoc, like Cross-Validation(CV), or not implementable in practice, like Lepski's method~\cite{Goldenshluger_2011, goldenshluger2013general}. Furthermore, in previous $\knn$ works, even after assuming equal weights, no proper attention has been given to theoretically justified data-driven tuning of $k$ without data-splitting/CV. Although NW regression prescribes adaptive weights $w$, tuning of $h$, and more importantly, choice of the kernel still lacks a principled approach.

Existing theory in $\knn$ demonstrates that the issue of properly choosing $k$ is inherently tied to the problem of carefully balancing the bias-variance decomposition of $\knn$ regression, where the variance of $\knn$ is largely determined by the underlying noise variance. This opens a possibility for automating the $k$ selection process by simultaneously tracking the noise variance along with fitting the data using $\knn$. 


The focus of our work is proposing a principled alternative to $\knn$ with automatic selection of $k$ and the corresponding weights for matrix completion with provable guarantees. We reformulate the mean squared error(MSE) of the classical weighted $\knn$ as a convex optimization problem over the weights given to the nearest neighbors. The corresponding optimal solution, coupled with a fixed point iteration over the noise variance, leads to the data-driven tuning mechanism. We further leverage the convex optimization step for establishing finite-sample guarantees (\cref{thm:mse_upper_boun_nonmissing}, \cref{thm:mse_general_bound}) with no assumptions on the underlying ground truth matrix, an unprecedented event in the literature of nearest neighbors in matrix completion.

\paragraph{Our Contributions}
\begin{enumerate}[label=(C\arabic*)]
    \item We propose the adaptively weighted nearest neighbors ($\awnn$): a weighted nearest neighbor based algorithm for matrix completion that does not rely on any heavy parametric assumption on the data generation process. 
    \item The requisite weights in $\awnn$ can be computed in closed form from the observed matrix itself. Because no iterative search or cross-validation is needed, the procedure not only runs quickly but also adapts the weights automatically to the data at hand. 
    \item We provide theoretical guarantees on the performance of $\awnn$
     without making any additional assumption (low rank/sparsity) on the structure of the ground truth matrix. To our knowledge, this is the first work on nearest neighbors to provide such a guarantee in the context of matrix completion. 
\end{enumerate}

\paragraph{Related Work} 

Matrix Completion algorithms are fundamentally of two types: Empirical Risk Minimization (ERM) methods and Nearest Neighbors (NN) methods. ERM methods~\cite{Chatterjee15, hastie2015matrix, bhattacharya2022matrix, xu2013speedup,jain2013provable, zhong2015efficient, chiang2015matrix,lu2016sparse, guo2017convex,eftekhari2018weighted, ghassemi2018global, chiang2018using,arkhangelsky2019synthetic,bertsimas2020fast, agarwal2020synthetic, agarwal2021causal} are historically more popular, with optimal guarantees for underlying matrix recovery under uniform missingess. But the theoretical guarantees and empirical performance of these methods generally break down when the missingness in the data matrix is confounded or is correlated to the underlying ground truth matrix. In contrast, the NN algorithms~\cite{dwivedi2022counterfactual, dwivedi2022doubly,  yu2022nonparametric, sadhukhan2024adaptivity} have been found to be robust to the missingness patterns of the matrix, and can yield desirable MSE rate even when some entries of the matrix are missing deterministically. Moreover under an implicit low-dimensional structure on the ground truth matrix, the NN algorithms tend to recover the minimax optimal non-parametric rate.

\paragraph{Organization}  
We discuss the problem set-up and motivate the idea behind the proposed algorithm in \Cref{sec:problem}. We introduce the $\awnn$ algorithm and study its theoretical guarantees in \Cref{sec:algo_theo}. We support the theoretical result with extentsive simulation study in \Cref{sec: simulation}. 

\section{Problem set-up and Motivation}
\label{sec:problem}
Our objective is to estimate the entries of the ground truth matrix $\Theta \in \mathbb{R}^{n \times m}$ (whose $(i,j)$-th entry is $\theta_{i,j}$) using the data matrix $X \in \mathbb{R}^{n \times m}$ whose $(i, j)$-th entry (for $(i,j) \in [n] \times [m]$) is given by, 
\begin{align}
    \label{eq:main_model}
    X_{i,j} = \begin{cases}
        \theta_{i,j} + \epsilon_{i,j} &\mbox{ if } A_{i,j} = 1, \\
        * & \mbox{ if } A_{i,j} = 0. 
    \end{cases} 
\end{align}
Here $\{A_{i,j}\}_{i \in [n], j \in [m]}$ are binary random variables denoting which entries of the matrix are observed. The noise terms $\epsilon_{i,j}$ are i.i.d.\ with variance $\sigma^2$. We propose the adaptively weighted row nearest neighbor algorithm ($\awnn$) in this section to estimate $\theta_{i,j}$ for $i \in [n], j \in [m]$. To explain the primary idea behind the algorithm we consider the simpler situation in this section when there is no missingness in the data i.e.\ $A_{i,j} = 1$ for all $i \in [n], j \in [m]$. Broadly speaking, this algorithm has two main steps: in the first step we obtain an estimate $\widehat \sigma^2$ of the error vaiance $\sigma^2$, in the second step we leverage the magnitude of the estimated error variance to construct a weighted nearest neighbors estimate $\widehat \theta_{i,j}  = \sum_{i' = 1}^n  w_{i'} X_{i',j} $ of $\theta_{i,j}$ where the weight $ w_{i'}$ on the entry from the $i'$-th row is inversely proportional to the ``distance" between the $i$-th row and $i'$-th row. 

We recall that the unweighted (vanilla) row nearest neighbor estimate (see \cite{dwivedi2022doubly}, \cite{sadhukhan2024adaptivity}) identifies a neighboring set of rows $\neighbors$ and estimates $\theta_{i,j}$ by taking average on the $j$-th column over all rows in $\neighbors$,
\begin{align}
\label{eq:vanilla_estimate}
    \widehat \theta_{i,j}^{\mathrm{UW}} = \sumn[i'] w_{i'} X_{i',j} \qtext{where} w_{i'} = \begin{cases}
        \frac{1}{|\neighbors|}& \qtext{if} i'\in \neighbors,\\
        0 &\qtext{otherwise}. 
    \end{cases}
\end{align}
However this does not provide ideal estimates in all situations. For instance if $\sigma^2 = 0$ and we have the prior knowledge of this fact, then the most natural estimate of $\theta_{i,j}$ is $X_{i,j}$. This suggests that we can use $\widehat \sigma^2$ to provide a more accurate estimate of the entries of the matrix by carefully weighing the neighboring rows provided $\widehat \sigma^2$ is a good enough estimate of $\sigma^2$. This is the main motivation behind the adaptively weighted row nearest neighbor algorithm. 

\subsection{Background and early insights}
To rigorously formulate the idea behind the proposed algorithm we consider the following notion of distances between the rows, 
\begin{align}
    \label{eq:oracle_row_dist}
    \rho_{i', i} = \frac{\sum_{j = 1}^m (\theta_{i,j} - \theta_{i', j})^2}{m} - 2\sigma^2 \quad \mbox{for all } i,i' \in [n]. 
\end{align}
We also assume that we have reliable estimates $\widehat \rho_{i', i}$ of the ``true" distances between the rows $\rho_{i', i}$. More formally, we make the following assumption.
\begin{assumption}[Concentration of NN distances]
    \label{assump:dist_concentration}
For any fixed $\delta \in (0,1)$ and for any pair of rows $i,i' \in [n]$ the event $\mathcal{E}_{dist}$
\begin{align}
\label{eq:estimated_d}
    |\widehat \rho_{i', i} - \rho_{i', i}| \leq  \zeta_{i',i}(m, \delta),
\end{align}
holds with probability at least $1 - \delta$.
\end{assumption}
We also make the following assumption on the noise terms to simplify the analysis.  
\begin{assumption}[Sub-gaussian noise]
  \label{assump:sub_g_noise}
The noise terms $\{\epsilon_{i,j}\}_{i \in [n], j \in [m]}$ are independent and identically distributed sub-gaussian random variables with mean $\mathbb{E}[\epsilon_{i,j}] = 0$ and $\mbox{Var}(\epsilon_{i,j}) = \sigma^2$ for all $i \in [n], j \in [m]$. 
\end{assumption}
The next theorem provides data-dependent bound on the row-wise mean squared error $(1/m)\sum_{j = 1}^m (\widehat \theta_{i,j} - \theta_{i,j})^2$ for the $i$-th row ($i \in [n]$) where $\widehat \theta_{i,j}  = \sum_{i' = 1}^n w_{i'} X_{i',j} $ and the weights $\{w_{i'}\}_{i' = 1}^n$ are non-negative and satisfy $\sum_{i'= 1}^n  w_{i'} = 1$. 
\begin{proposition}
    \label{thm:row_wise_bounds}
Let $\widehat \theta_{i,j}  = \sum_{i' = 1}^n w_{i'} X_{i',j} $ where $ w^{(i)} = ( w_{1}, \cdots, w_{n})$ is the weight vector associated with the $i$-th row. Under assumptions~\ref{assump:dist_concentration} and \ref{assump:sub_g_noise}, the row-wise mean squared error for the $i$-th row ($i \in [n]$) has the following data-dependent upper bound with probability at-least $1 - 3\delta$, 
\begin{align}
    \frac{1}{m}\sum_{j = 1}^m (\widehat \theta_{i,j} - \theta_{i,j})^2 \leq 2 \left\{2\log(2m/\delta)\sigma^2\| w^{(i)}\|_2^2 + \sum_{i' = 1}^n  w_{i'} \widehat \rho_{i', i}   \right\} + 2 \max_{i' \in [n]} \zeta_{i',i}(m , \delta). 
\end{align}
\end{proposition}
The proof of \Cref{thm:row_wise_bounds} is discussed in \Cref{app:proof_row_wise_bounds}. The second term in the upper bound on the row MSE, $\max_{i' \in [n]} \zeta_{i',i}(m , \delta)$ captures the hardness in estimating the ``true" distances $\rho_{i', i}$ between the rows and does not depend on how we weigh the rows to estimate the entries of the matrix. Therefore \Cref{thm:row_wise_bounds} suggests that we can optimize the row-wise MSE by choosing weights $\widehat w^{(i)} = (\widehat w_{1}, \cdots, \widehat w_{n})$ that minimize the first term of the upper bound on the row MSE. During implementation, we replace $\sigma^2$ by its consistent estimate $\widehat \sigma^2$. We state the steps of the adaptively weighted row nearest neighbor algorithm ($\awnn$) below when the matrix $X$ comprises of no missing entries.
\begin{enumerate}
    \item Obtain a consistent estimate $\widehat \sigma^2$ of the variance of the noise terms. 
    \item Compute the optimal weight vector $\widehat w^{(i)} = (\widehat w_{1}, \cdots, \widehat w_{n})$ that minimizes the following objective function,
    \begin{align}
    \label{eq:algo_step_nomissing}
  \widehat w^{(i)}    \coloneqq\mbox{arg min}_{w^{(i)}}\brackets{2\log(2m/\delta)\widehat \sigma^2\| w^{(i)}\|_2^2+ \sum_{i' = 1}^n  w_{i'} \widehat \rho_{i', i} },
\end{align}
where $w^{(i)} = (w_1, \cdots, w_n)$ is used to denote a candidate weight vector i.e.\ $w^{(i)}$ is an elemeny-wise non-negative vector that satisfies $\sum_{i' = 1}^n w_{i'} = 1$.
    \item Return the weighted nearest neighbor estimate, 
    \[
    \widehat \theta_{i,j} = \sum_{i' = 1}^n \widehat w_{i'} X_{i',j}. 
    \]
\end{enumerate}
 For showing the theoretical guarantees in this paper we assume that $\sigma^2$ is known a priori for ease of explanation. All the derivations can be re-produced when do not know the true variance but have a consistent estimate $\widehat \sigma^2$ of $\sigma^2$. Returning back to the discussion of the noiseless setting ($\sigma^2 = 0$), we observe that our proposed weight vector $ \widehat w^{(i)} $ solves the following optimization problem, 
\begin{align}
\label{eq:opt_noiseless}
    \widehat w^{(i)} = \mbox{arg min}_{w^{(i)}} \brackets{\sum_{i' = 1}^n  w_{i'} \widehat \rho_{i', i} }. 
\end{align}
Since $\widehat \rho_{i', i}= 0$, it is easy to check that the objective function is minimzed at $\widehat w_i = 1$ and $\widehat w_{i'} = 0$ for all rows $i' \neq i$ and the minimum value of the objective function is $0$. Hence for all $j \in [m]$, this algorithm yields the estimate $\widehat \theta_{i,j} = \sum_{i' = 1}^n \widehat w_{i'}X_{i',j} = X_{i,j} $ which is exactly same as $\theta_{i,j}$ (as $\epsilon_{i,j} = 0$). This demonstrates that, in the idealized noiseless setting, the adaptively weighted row nearest neighbor approach yields more accurate matrix entry estimates than the standard row nearest neighbor method. 

Even when self-neighbor is not allowed by explicitly setting $\widehat w_i = 0$ in the above optimization, the adaptively weighted nearest-neighbor method still delivers more accurate estimates of the ground-truth matrix in the noiseless setting. To see this, we assume that the entries of the ground truth matrix come from a low rank factor model. In particular, we assume that $\theta_{i, j} = \langle u_i, v_j \rangle$ for all $(i , j) \in [n] \times [m]$ where $u_1, \cdots, u_n $ are iid $d$-dimensional row latent factors with covariance matrix $\mathbb{E}u_1u_1^T = I_d$ and $v_1, \cdots, v_n $ are iid $d$-dimensional column latent factors with covariance matrix $\mathbb{E}v_1v_1^T = I_d$. In noiseless setting, the estimated distance between the rows $i, i'$, 
\begin{align}
\label{eq:simplified_model_dist}
    \widehat \rho_{i', i} = \rho_{i', i}  = \frac{1}{m} \sum_{j = 1}^m \langle u_i - u_{i'} ,v_j \rangle  = (u_i - u_{i'})^T \left(\frac{1}{m} \sum_{j = 1}^m v_j v_j^T \right) (u_i - u_{i'}) = \|u_i - u_{i'}\|_2^2 + o_P(1). 
\end{align}
If self-neighbor is not allowed, the objective function in \eqref{eq:opt_noiseless} is minimized at $\widehat w_{i_{(1)}} = 1$ and $\widehat w_{i'} = 0$ for all rows $ i' \neq i_{(1)}$ where $i_{(1)}$ is the closest row to the $i$-th row i.e.\ $\widehat \rho_{i, i_{(1)}} = \min_{i' \neq i}\widehat \rho_{i', i}$. Observe that by \eqref{eq:simplified_model_dist}, the index $i_{(1)}$ corresponds to the row whose latent factor $u_{i_{(1)}}$ is nearest in Euclidean norm to the latent factor $u_i$ of the $i$-th row.  Therefore when self-neighbor is not allowed, the adaptively weighted nearest-neighbor method estimates $\theta_{i,j}$ by $\theta_{i_{(1)}, j}$ where $\|u_i - u_{i_{(1)}}\|_2 = \min_{i' \neq i } \| u_i - u_{i'}\|_2$. It can be easily checked that this is the best one can do without any other additional information. We note that by cauchy-schwartz inequality, $|\sum_{i' \neq i} w_{i'}\theta_{i',j} - \theta_{i,j}| = |\langle \sum_{i' \neq i}w_{i'}(u_{i'} - u_i), v_j \rangle| \leq \|v_j\|_2\sum_{i' \neq i} w_{i'} \|u_{i'} - u_i \|_2 $. Consequently, the weight should be allocated exclusively to the row whose latent factor lies nearest to $u_i$. 

\section{Algorithm and Theoretical guarantees}
\label{sec:algo_theo}
This section presents the $\awnn$ algorithm for the general model \eqref{eq:main_model}, followed by a performance analysis across various regimes.

\begin{algorithm2e}[H]
\caption{$\awnn$\ --\ Return estimates of counterfactual signals $\theta_{i,j}$ in a matrix of size $n\times m$ with small MSE} 
  \label{algo:row_star_NN}
  \SetAlgoLined
  \DontPrintSemicolon
  \small
  {
\KwIn{\textup{$n\times m$ Data Matrix $X$, probability $\delta$}}
  \BlankLine
    Initialize $\what{\sigma}^2\gets \Vararg{\parenth{X_{i,j}}_{i,j\in [n]\times[m]} }/10$\\
    \BlankLine
    $\drown{i'}{i} \gets$ L2-row-distance$\parenth{X_{i'\cdot},X_{i\cdot}}$, $\forall i',i \in[n]$ s.t. $i'\neq i$\\[2pt]
    \BlankLine
    $\parenth{\what\theta_{i,j}}_{i,j\in [n]\times[m]} \gets$ {\normalsize\rowimputelink}\,$\parenth{X, \what\sigma^2, \delta,\parenth{\drown{i'}{i}}_{i',i\in[n]^2}}$
    \\[2pt]
    \BlankLine
    $\what{\sigma}^2\gets \frac{1}{nm}\sum_{i\in[n],j\in[m]}\parenth{X_{i,j}-\what{\theta}_{i,j}}^2$\\
    \BlankLine
    Repeat Step 3 till the difference between new and old $\what\sigma^2$ is small.\\
    \BlankLine
}
  \KwOut{\textup{$\parenth{\what\theta_{i,j}}_{i,j\in [n]\times[m]}$}}
\end{algorithm2e}

\begin{algorithm2e}[H]
\caption{Row Imputer\ --\ Return weighted average of $\parenth{X_{i,j}}_{i,j\in [n]\times[m]}$ as the estimates of $\theta_{i,j}$'s} 
  \label{algo:row_imputer}
  \SetAlgoLined
  \DontPrintSemicolon
\KwIn{\textup{$n\times m$ Data Matrix $X$, noise variance estimate $\what\sigma^2$, probability $\delta$, row-wise distances $\what{\rho}_{\cdot, \cdot}$ }}
  {
 \BlankLine
$\what{\rho}_{\cdot, \cdot} \gets \what{\rho}_{\cdot, \cdot} - 2\what\sigma^2, \mathrm{Diag}\parenth{\what{\rho}_{\cdot, \cdot}}\gets 0$\\[2pt]
    \For{$j = 1,\ldots, m$}
        {
        $\mathcal{A}(j)\gets\braces{i'\in [n]|A_{i',j} = 1}$ 
            \\[2pt]
        \For{$i = 1, \ldots, n$}
            {
            $u \gets $ {\normalsize \wgtadjlink}$\parenth{\braces{\frac{1}{\abss{\mathcal{A}(j)}}-\frac{1}{4\what\sigma^2\log(1/\delta)}\parenth{\drown{i'}{i} - \frac{1}{|\mathcal{A}(j)|} \sum_{i'' \in \mathcal{A}(j)}\drown{i''}{i}}}_{i'\in[n]}}$\\[2pt]
            $\widehat w_{i',j}(i) \gets$ $\begin{cases}
                &\max\braces{0, \frac{1}{\abss{\mathcal{A}(j)}}-\frac{1}{4\what\sigma^2\log(1/\delta)}\parenth{\drown{i'}{i} - \frac{1}{|\mathcal{A}(j)|} \sum_{i'' \in \mathcal{A}(j)}\drown{i''}{i}} - u}, \text{ if } i'\in \mathcal{A}(j)\\
                &0\hspace{10cm}\qtext{else}
            \end{cases}$\\[2pt]
            $\what\theta_{i,j}\gets\sum_{i\in\mathcal{A}_j}\widehat w_{i',j}(i) X_{i,j}$
            }
        }      
}
\KwOut{\textup{estimated signal matrix $\parenth{\what\theta_{i,j}}_{i,j\in [n]\times[m]}$}}
\end{algorithm2e}

\begin{algorithm2e}[H]
\caption{Weight Adjuster\ --\ For a sequence $\pseqann$ with $\sumn a_i=1$, returns $u$ such that $\sumn\max\braces{a_i-u,0}=1$} 
  \label{algo:wgt_adjuster}
  \SetAlgoLined
  \DontPrintSemicolon
\KwIn{\textup{$n$-length sequence $\pseqann$ such that $\sumn a_i=1$}}
  \small
  {
  \BlankLine
    $\pseqnn{b}{n}\gets \text{Sort}\parenth{\pseqann} \text{ in decreasing order}$\\[2pt]
    \BlankLine
    Initialize $\text{sum}\gets 0$\\[2pt]
    \BlankLine
    \For{$k=1,\ldots, n$}
        {
        $\text{sum}\gets\text{sum} + b[k]$\\[2pt]
        $u \gets \parenth{\text{sum} - 1}/k$\\[2pt]
        \If{$k == n \qtext{or} b_{k+1}\leq u$}
            {break}
        }
}
\KwOut{\textup{$u$}}
\end{algorithm2e} 
\subsection{Matrix denoising}
\label{subsec: matrix completion}
In this subsection we study the performance of algorithm~\ref{algo:row_star_NN} under the assumption that there is no missingness in the data. The algorithm~\ref{algo:row_star_NN} boils down to the optimization problem \eqref{eq:algo_step_nomissing} when there are no missing entries in the data. It can be shown that the optimization problem in \eqref{eq:algo_step_nomissing} effectively picks a row-neighbor set $\mathcal{R}(i)$, assigning zero weight to every row outside it. The precise statement is given by the theorem that follows.
\begin{proposition}
    \label{thm:optimal_weights}
Suppose the weight vector $\widehat w^{(i)} = (\widehat w_{1}, \cdots, \widehat w_{n})$ is the solution of the optimization problem in \eqref{eq:algo_step_nomissing}. Then there exists a set of neighboring rows $\mathcal{R}(i) \subset [n]$ of cardinality $K_i = | \mathcal{R}(i)|$ such that, 
\begin{align}
     \what{w}_{i'} &= \indic{i'\in\mathcal{R}(i)}\brackets{\frac{1}{K_{i}} - \frac{1}{4 \log(2m/\delta)\sigma^2}\parenth{\drown{i'}{i} - \Bar{\widehat \rho}_i}} \quad \mbox{for all} \quad i' \in [n], 
\end{align}
where $\Bar{\widehat \rho}_i = (1/ K_i)\sum_{i' \in \mathcal{R}(i)} \drown{i'}{i}$ is the mean of the estimated distances between the $i$-th row and the rows in the set $\mathcal{R}(i)$. 
\end{proposition}
\begin{remark}
\Cref{thm:optimal_weights} establishes that in order to estimate $\theta_{i,j}$, $\awnn$ confines the positive weights to a proximity set $\mathcal{R}(i)$ and within that set, gives higher weight to any row that lies closer to row $i$. In other words, for any two rows $i', i'' \in \mathcal{R}(i)$ such that $\drown{i'}{i} \leq \drown{i''}{i}$ we have $\widehat w_{i'} \geq \widehat w_{i''}$, meaning rows nearer to $i$ receive larger weight in estimating $\theta_{i,j}$. 
\end{remark}
For the detailed proof of \Cref{thm:optimal_weights}, refer to \Cref{app:proof_optimal_weights}. Together, the optimal-weight characterization (\Cref{thm:optimal_weights}) and the row-wise MSE bound (\Cref{thm:row_wise_bounds}) provide a data dependent upper bound on each row’s mean-squared error.
\begin{theorem}
    \label{thm:mse_upper_boun_nonmissing}
Suppose $\widehat w^{(i)} = (\widehat w_{1}, \cdots, \widehat w_{n})$ is the optimal weight vector selected by $\awnn$. Under assumptions~\ref{assump:dist_concentration} and \ref{assump:sub_g_noise}, the row-wise mean squared error for the $i$-th row ($i \in [n]$) has the following data-dependent upper bound with probability at-least $1 - 3\delta$, 
\begin{align}
   & \frac{1}{m}\sum_{j = 1}^m (\widehat \theta_{i,j} - \theta_{i,j})^2 \leq  \mathbb{B}_i^{\mathrm{AW}} + \mathbb{V}_i^{\mathrm{AW}} \quad \mbox{where},\\
    &  \mathbb{B}_i^{\mathrm{AW}} = 2 \left\{\Bar{\widehat \rho}_i - \frac{\sum_{i' \in \mathcal{R}(i)} \left(\drown{i'}{i} - \Bar{\widehat \rho}_i\right)^2}{8\log(2m/\delta)\sigma^2} + \max_{i' \in [n]} \zeta_{i',i}(m , \delta) \right\}, \\
    & \mathbb{V}_i^{\mathrm{AW}}=  \frac{4\log(2m/\delta)\sigma^2}{K_i}. 
\end{align}
\end{theorem}
See \Cref{app:proof_upper_bound_non_missing} for the proof of \Cref{thm:mse_upper_boun_nonmissing}. \Cref{thm:mse_upper_boun_nonmissing} provides a general error bound on the $\awnn$ estimates and decomposes the error bound into a bias term $\mathbb{B}_i^{\mathrm{AW}}$ and a variance term $\mathbb{V}_i^{\mathrm{AW}}$. In order to interpret the error bound, let $\eta_i^2 = \max\{\drown{i'}{i}: i' \in \mathcal{R}(i)\}$ denote the distance of $i$-th row from the farthest row in the proximity set $\mathcal{R}(i)$. Using Von Szokefalvi Nagy inequality (see \cite{kaiblinger2020inequality}) we can show that $\sum_{i' \in \mathcal{R}(i)} \left(\drown{i'}{i} - \Bar{\widehat \rho}_i\right)^2 \geq \eta_i^4/2$. Consequently, the bias component of the error bound is no greater than,
\begin{align}
\label{eq:bias_awnn_upper_bound}
   \mathbb{B}_i^{\mathrm{AW}} \leq 2 \left\{\eta_i^2 - \frac{\eta_i^4}{16\log(2m/\delta)\sigma^2} + \max_{i' \in [n]} \zeta_{i',i}(m , \delta)\right\}. 
\end{align}
It can be shown that this is an improvement on the bias in the error bound of the unweighted row nearest neighbor estimate \eqref{eq:vanilla_estimate} that selects a set $\neighbors(\eta^2) = \{i' \in [n]: \drown{i'}{i} \leq   \eta^2    \}$ of nearest neighbors of row $i$. Let $k_{\eta^2} = |\neighbors(\eta^2)|$ be the cardinality of the nearest neighbor set i.e.\ $\eta^2$ denotes the distance $\drown{i'}{i}$ of the $k_{\eta^2}$-th farthest row $i'$ from the row $i$. Under assumptions~\ref{assump:dist_concentration} and \ref{assump:sub_g_noise}, the row-wise mean squared error of the unweighted row nearest neighbor estimate (that thresholds at $\eta^2$) for the $i$-th row ($i \in [n]$) has the following upper bound with probability at-least $1 - 2\delta$,
\begin{align}
\label{eq:mse_decomp_vanilla}
   & \frac{1}{m}\sum_{j = 1}^m (\widehat \theta_{i,j}^{\mathrm{UW}} - \theta_{i,j})^2 \leq  \mathbb{B}_i^{\mathrm{UW}}(\eta^2) + \mathbb{V}_i^{\mathrm{UW}}(\eta^2) \quad \mbox{where},\\
    &  \mathbb{B}_i^{\mathrm{UW}}(\eta^2) = 2 \left\{\frac{\sum_{i' \in \neighbors(\eta^2)}\drown{i'}{i}}{k_{\eta^2}} + \max_{i' \in [n]} \zeta_{i',i}(m , \delta) \right\}, \\
    & \mathbb{V}_i^{\mathrm{UW}}(\eta^2) =  \frac{4\log(2m/\delta)\sigma^2}{k_{\eta^2}}. 
\end{align}
See \cite{dwivedi2022counterfactual} for a proof of the above statement. It is clear from this decomposition that for $\eta^2 = \eta_i^2$, the bias $\mathbb{B}_i^{\mathrm{UW}}(\eta_i^2)$ of the unweighted row nearest neighbor estimate is less than the bias $ \mathbb{B}_i^{\mathrm{AW}}$ of the $\awnn$ estimate because of the additional debiasing term $\eta_i^4/(16\log(2/\delta)\sigma^2)$. More generally we can show the following lemma. 
\begin{lemma}
    \label{lem:bias_comp}
The $\awnn$ algorithm guarantees a lower row wise mean squared error than the best-performing unweighted row-nearest-neighbour estimator, even when the latter’s search radius $\eta^2$ is optimally tuned. 
\begin{align}
    \mathbb{B}_i^{\mathrm{AW}} + \mathbb{V}_i^{\mathrm{AW}} \leq \min_{\eta >0}\{\mathbb{B}_i^{\mathrm{UW}}(\eta^2) + \mathbb{V}_i^{\mathrm{UW}}(\eta^2) \}. 
\end{align}
\end{lemma}
The proof of \Cref{lem:bias_comp} is illustrated in \Cref{app:bias_comp}. \Cref{lem:bias_comp} shows that, by appropriately weighting rows within a selected neighborhood, $\awnn$ achieves a tighter row-wise mean-squared error bound than the unweighted row nearest neighbor method. To further unpack the statements in \Cref{thm:mse_upper_boun_nonmissing} and \Cref{lem:bias_comp} we need additional conditions on the ground truth matrix. In particular, we assume that $\Theta$ has an implicit low-dimensional structure. 
\begin{assumption}
    \label{assump:low_dimensional}[Non-linear factor model]
Conditioned on the row latent factors $u_1, \cdots, u_n \in \mathbb{R}^{d_1}$ and the column latent factors $v_1, \cdots, v_m \in \mathbb{R}^{d_2}$, the ground truth has the following low-rank representation, 
\begin{align}
    \theta_{i,j} = f(u_i, v_j) \quad \forall \quad (i,j) \in [n] \times [m],
\end{align}
where $f$ is $(\lambda, L)$ H\"older function for $\lambda \in (0,1]$ i.e.\ for $x, x' \in \mbox{Domain}(f)$, 
\begin{align}
    |f(x) - f(x')| \leq L\|x - x'\|_{\infty}^{\lambda}. 
\end{align}
\end{assumption}
For ease of presentation we also make assumption on the sampling of the row latent factors. These type of assumptions are not uncommon in the literature (see \cite{yu2022nonparametric, sadhukhan2024adaptivity}). 
\begin{assumption}[Row latent factors]
\label{assump:row_latent}
    The row latent factors $u_1, \cdots, u_n$ are sampled independently from $\mathrm{Uniform}[0,1]^{d_1}$.
\end{assumption}
Choosing the unit hypercube as the support of the latent factors is not restrictive: the same theory holds for any compact support in $\mathbb{R}^{d_1}$. The proofs generalize to arbitrary sampling laws once the uniform tail bounds are replaced by those of the chosen distribution.
\begin{corollary}
\label{cor:exact_mse_rate}
    Under assumptions~\ref{assump:dist_concentration}, \ref{assump:sub_g_noise}, \ref{assump:low_dimensional}, and \ref{assump:row_latent}, $\awnn$ achieves the following rate with probability at-least $ 1- 4\delta$,
    \begin{align}
        \frac{1}{m}\sum_{j = 1}^m (\widehat \theta_{i,j} - \theta_{i,j})^2 = \begin{cases}
          O \left(n^{-\frac{2\lambda}{d_1 + 2\lambda}} \right) \quad & \mbox{if}\quad n = O\left(\{ \max_{i' \in [n]} \zeta_{i',i}(m , \delta)\}^{-\frac{d_1 + 2\lambda}{2 \lambda}}\right)  , \\
          O\left(\max_{i' \in [n]} \zeta_{i',i}(m , \delta) \right) \quad & \mbox{otherwise}. 
        \end{cases}
    \end{align}
\end{corollary}
\Cref{cor:exact_mse_rate} is a consequence of \Cref{thm:mse_upper_boun_nonmissing} and \Cref{lem:bias_comp}. See \Cref{app:exact_mse_rate_non_missing} for the detailed proof. The term $n^{-2\lambda/(d_1 + 2\lambda)}$ is exactly the convergence rate attained by the unweighted row nearest neighbour estimator when its radius $\eta$ is optimally tuned. Consequently, \Cref{cor:exact_mse_rate} shows that in the regime $n = O(\{ \max_{i' \in [n]} \zeta_{i',i}(m , \delta)\}^{-(d_1 + 2\lambda)/2\lambda})$, the $\awnn$ estimate attains the same row MSE decay rate without the burden of tuning the radius parameter $\eta$. The next subsection presents the generalized $\awnn$ algorithm for handling missing entries in the observed matrix.
\subsection{Matrix completion}
\label{subsec: matrix completion}
When there is missingness in the data, we can not select a single proximity set $\mathcal{R}(i)$ for the $i$-th row and estimate $\theta_{i,j}$ (for any $j\in [m]$) by averaging over the entries in the $j$-th column of all the rows in the set $\mathcal{R}(i)$. This difficulty arises because rows within the selected neighborhood $\mathcal{R}(i)$ typically contain multiple missing entries. To address this issue, when there are missing entries in the observed matrix, we do not use the same weighing to estimate all the entries in a given row. The algorithm~\ref{algo:row_star_NN} can be restated as follows when there is missingness in the data.  
\begin{enumerate}
    \item Obtain a consistent estimate $\widehat \sigma^2$ of the variance of the noise terms. 
    \item Identify the active set $\mathcal{A}(j)$ of rows for the $j$-th column, 
    \begin{align}
        \mathcal{A}(j) = \{ i \in [n]: A_{i,j} = 1\}.
    \end{align}
   \item Compute the optimal weight vector $\widehat w^{(i, j)} = (\widehat w_{1, j}, \cdots, \widehat w_{n, j})$ that minimizes the following objective function,
    \begin{align}
    \label{eq:algo_step_missing}
  \widehat w^{(i, j)}    \coloneqq\mbox{arg min}_{w^{(i, j)}}\brackets{2\log(2m/\delta)\widehat \sigma^2\|w^{(i,j)} \|_2^2 + \sum_{i' \in \mathcal{A}(j)}  w_{i', j} \widehat \rho_{i', i} },
\end{align}
where $w^{(i, j)} = (w_{1, j}, \cdots, w_{n, j})$ is used to denote a candidate weight vector i.e.\ $w^{(i, j)}$ is an elemeny-wise non-negative vector that satisfies $\sum_{i' = 1}^n w_{i', j}A_{i',j} = 1$.
    \item Return the weighted nearest neighbor estimate, 
    \[
    \widehat \theta_{i,j} = \sum_{i' = 1}^n \widehat w_{i', j} A_{i',j}X_{i',j}. 
    \]
\end{enumerate}
As in the previous sub-section, to simplify the discussion we show the theoretical guarantees in this sub-section assuming known $\sigma^2$. From \Cref{thm:optimal_weights} we know that the optimal weight vector $\widehat w^{(i,j)}$ that minimizes the objective function in \eqref{eq:algo_step_missing} has the alternative representation, 
\begin{align}
 \label{eq: weight defn}
    \widehat w_{i',j} = \indic{i'\in\mathcal{R}(i, j)}\brackets{\frac{1}{K_{i, j}} - \frac{1}{4 \log(2m/\delta)\sigma^2}\parenth{\drown{i'}{i} - \Bar{\widehat \rho}_{i;j}}} \quad \mbox{for all} \quad i' \in \mathcal{A}(j),
\end{align}
where $\mathcal{R}(i,j) \subset \mathcal{A}(j)$ is a set of nearest neighbors of cardinality $K_{i,j}$ automatically identified by $\awnn$ and $\Bar{\widehat \rho}_{i;j} = (1/ K_{i,j})\sum_{i' \in \mathcal{R}(i,j)} \drown{i'}{i}$ is the mean of the estimated distances of the rows in the set $\mathcal{R}(i, j)$ from the $i$-th row. Note that higher the value of $\drown{i'}{i}$, the lower the value of $\what{w}_{i',j}$. A salient feature of the $\awnn$ method is that it by design excludes rows distant from row $i$ when constructing the neighbourhood $\mathcal{R}(i,j)$, thereby enhancing the reliability of the resulting matrix‐entry estimates. The next lemma presents a formal statement of this property.
\begin{lemma}
\label{lem:distance_bound}
    In estimating ${\theta}_{i,j}$, any row $ i' $ in the proximity set $\NR[j]$ satisfies,
    \begin{align}
        \drown{i'}{i} \leq \begin{cases}
            4 \log(2m\delta)\sigma^2,& \qtext{if $\miss[i]=1$,}\\
            4 \log(2m\delta)\sigma^2 + \min_{i' \in \NR[j]}\drown{i'}{i},& \qtext{if $\miss[i]=0$.}
        \end{cases} 
    \end{align}
    Additionally the average of the empirical distances over the proximity set $\NR[j]$ satisfy,
    \begin{align}
        \drownmean{j} = \begin{cases}
        4 \log(2m\delta)\sigma^2\parenth{\infnorm{\what{w}^{(i, j)}} - \frac{1}{\K[j]}},& \qtext{if $\miss[i]=1$,}\\
            4 \log(2m\delta)\sigma^2\parenth{\infnorm{\what{w}^{(i, j)}} - \frac{1}{\K[j]}} + \min_{i' \in \NR[j]}\drown{i'}{i},& \qtext{if $\miss[i]=0$.}
        \end{cases} 
    \end{align}
\end{lemma}
The proof of \Cref{lem:distance_bound} uses the representation \eqref{eq: weight defn} and the fact that $\widehat w_{i', j}$ lies in $[0,1]$ for all $i' \in [n], j \in [m]$. For the complete proof, see \Cref{app:distance_bound_awnn}. \Cref{lem:distance_bound} establishes that if a row $i'$ belongs to the nearest neighbor set $\NR$ of the $i$-th row, its estimated distance $\drown{i'}{i}$ from the $i$-th row can not be larger than $4 \log(2m\delta)\sigma^2$ when $A_{i,j}=1$. The following theorem discusses the performance of $\awnn$ in the general setup. 
\begin{theorem}
    \label{thm:mse_general_bound}
Suppose $\widehat w^{(i,j)} = (\widehat w_1, \cdots, \widehat w_n)$ is the optimal weight vector selected by $\awnn$ to estimate $\theta_{i,j}$ in the presence of missing observations. Under assumptions~\ref{assump:dist_concentration} and \ref{assump:sub_g_noise}, the row-wise mean squared error for the $i$-th row ($i \in [n]$) has the following data-dependent upper bound with probability at-least $ 1- 4\delta$, 
\begin{align}
   & \frac{1}{m}\sum_{j = 1}^m (\widehat \theta_{i,j} - \theta_{i,j})^2 \leq  \mathbb{B}_i^{\mathrm{AW}} + \mathbb{V}_i^{\mathrm{AW}} \quad \mbox{where},\\
    &  \mathbb{B}_i^{\mathrm{AW}} = \frac{2}{m}  \sum_{j = 1}^m \drownmean{j} +  \frac{2}{m}  \sum_{j = 1}^m  \sum_{i' \in \NR} \widehat w_{i',j} \left(16M^2 \sqrt{\log(2/\delta)} (\|\widehat w_{i',\cdot}\|_2/ \|\widehat w_{i',\cdot}\|_1) + \zeta_{i', i}(m , \delta) \right) , \\
    & \mathbb{V}_i^{\mathrm{AW}}= \frac{4\log(2m/\delta)\sigma^2}{m}\sum_{j=1}^m \frac{1}{\K[j]}. 
\end{align}
\end{theorem}
The proof of \Cref{thm:mse_general_bound} is discussed in \Cref{app:mse_general_bound}. \Cref{thm:mse_general_bound} generalizes \Cref{thm:mse_upper_boun_nonmissing} to cover arbitrary missingness patterns. The second component in the bias term $\mathbb{B}_i^{\mathrm{AW}}$ characterizes the hardness in estimating the true distances $\rho_{i',i}$ by $\drown{i'}{i}$ and is generally $o((1/m)\sum_{j = 1}^m \drownmean{j})$. For the sake of clarity, we adopt the following assumption for the remainder of the paper, 
\begin{align}
\label{eq:clarity_assump}
   \frac{2}{m}  \sum_{j = 1}^m  \sum_{i' \in \NR} \widehat w_{i',j} \left(16M^2 \sqrt{\log(2/\delta)} (\|\widehat w_{i',\cdot}\|_2/ \|\widehat w_{i',\cdot}\|_1) + \zeta_{i', i}(m , \delta) \right) = o\left(  \frac{2}{m}  \sum_{j = 1}^m \drownmean{j}\right). 
\end{align}
Analogous to the preceding subsection, we adopt assumptions~\ref{assump:low_dimensional} and \ref{assump:row_latent} for the true matrix so that the upper bound can be interpreted more transparently. Moreover, for ease of explanation we make a missing completely at random (MCAR) assumption.
\begin{assumption}[MCAR missingness]
\label{asump_missingness}The indicators  $A_{i, j}$ are drawn \iid $\trm{Ber}(p)$, and independently of the latent factors and the noise. 
\end{assumption}
Invoking Assumptions \ref{assump:low_dimensional}, \ref{assump:row_latent}, and \eqref{eq:clarity_assump} and unpacking \Cref{thm:mse_general_bound} via the techniques of \Cref{lem:bias_comp} and \Cref{cor:exact_mse_rate}, we derive the following bound, which holds with probability no less than $1 - 4\delta$,
\begin{align}
 \frac{1}{m}\sum_{j = 1}^m (\widehat \theta_{i,j} - \theta_{i,j})^2 \leq &   ( 1+ o(1))\left(\frac{2}{m}  \sum_{j = 1}^m \drownmean{j} \right) +  \frac{4\log(2m/\delta)\sigma^2}{m}\sum_{j=1}^m \frac{1}{\K[j]} \\
 \leq &  \min_{\eta >0} \left\{ 2 \eta^2( 1+ o(1)) + \frac{4\log(2m/\delta)\sigma^2}{(1 - \delta)n p\left(\frac{\eta^2(1 - o(1))}{L^2} \right)^{\frac{d_1}{2\lambda}}}  \right\}.
\end{align}
One can easily check that the above expression is optimized for $\eta = O(n^{-\lambda/(d_1 + 2\lambda)}) $. The following corollary summarizes this result. 
\begin{corollary}
    \label{cor:gen_missing_MCAR}
Under assumptions~\ref{assump:dist_concentration}, \ref{assump:sub_g_noise}, \ref{assump:low_dimensional}, \ref{assump:row_latent}, and \eqref{eq:clarity_assump} $\awnn$ achieves the following row-wise mean squared error rate with probability at-least $ 1- 4\delta$,
    \begin{align}
        \frac{1}{m}\sum_{j = 1}^m (\widehat \theta_{i,j} - \theta_{i,j})^2 =  O \left(n^{-\frac{2\lambda}{d_1 + 2\lambda}} \right). 
    \end{align}    
\end{corollary}
\Cref{cor:gen_missing_MCAR} confirms that under MCAR missingness, $\awnn$ delivers the same rate as the best unweighted row-nearest-neighbour estimator while bypassing any need for hyper-parameter selection or cross-validation. We note that our focus on MCAR is solely for exposition; \Cref{thm:mse_general_bound} can be applied without modification to missing not at random (MNAR) regimes.

\section{Simulation Study}
\label{sec: simulation}

We demonstrate the practical usability of $\awnn$ by inspecting its empirical performance on synthetic datasets. We generate the data from model \cref{eq:main_model}, where the signals are created using a $(\lambda,2)$ \Holder function $f:\real^{2d}\to \real$ 
\begin{align}
\label{eq: simulation f}
    f\parenth{u_i,v_j} \coloneq \sum_{k=1}^d\abss{u_{i,k}+v_{j,k}}^\lambda \mrm{Sgn}\parenth{u_{i,k}+v_{j,k}}
\end{align}

\noindent with $\lambda\in[0,1]$, and the $d$ dimensional latent variables are generated as follows
\begin{align}
    u_i\sim \Unif[-0.5, 0.5]^d;&\quad v_j\sim \Unif[-0.5, 0.5]^d;\\
   \theta_{i,j} = f(u_i, v_j);&\quad \eps\sim\Gsn(0, \sigma_{\eps}^2),
\end{align}

We considered two types of missingness, a setup where all entries are observed and MCAR missingness setup $\parenth{p = 0.65, \cref{asump_missingness}}$. To improve interpretability and reproducibility, we present our findings by fixing Signal-to-Noise Ratio $(\msf{SNR})$ and choosing $\sigma_\eps^2$ accordingly. By default, we set the latent variable dimension $d = 2$ and $\msf{SNR} = 1$.

\begin{align}
    \msf{SNR} = 
    \sqrt{\frac{\sum_{i = 1}^n\sum_{j = 1}^m f^2(u_i, v_j)}{mn \sigma_{\eps}^2}}.
\end{align}

\paragraph{Baselines} We compare $\awnn$ with the unweighted row nearest neighbors (RowNN) \cite{li2019nearest, dwivedi2022counterfactual} to illustrate empirical benefits of weights in the nearest neighbor approach. We also added Universal Singular Value Thresholding (USVT) \cite{Chatterjee15} to the benchmark methods as a conventional matrix completion procedure. Lastly, we evaluate the Oracle $\awnn$ ($\oawnn$), which knows the true value of noise variance $\sigma_\eps^2$. $\oawnn$ directly uses the $\sigma_\eps^2$ to compute the weights assigned to the nearest neighbors instead of doing the fixed point iteration over $\what\sigma_\eps^2$ in $\rownnlink$.

\begin{figure*}[h]
    \centering
    \begin{tabular}{cc}
         \quad\qquad \textbf{No Missingness} & \quad \qquad \textbf{MCAR} \\
        \includegraphics[trim=0in 0in 0in 0in, clip, width=0.47\textwidth]{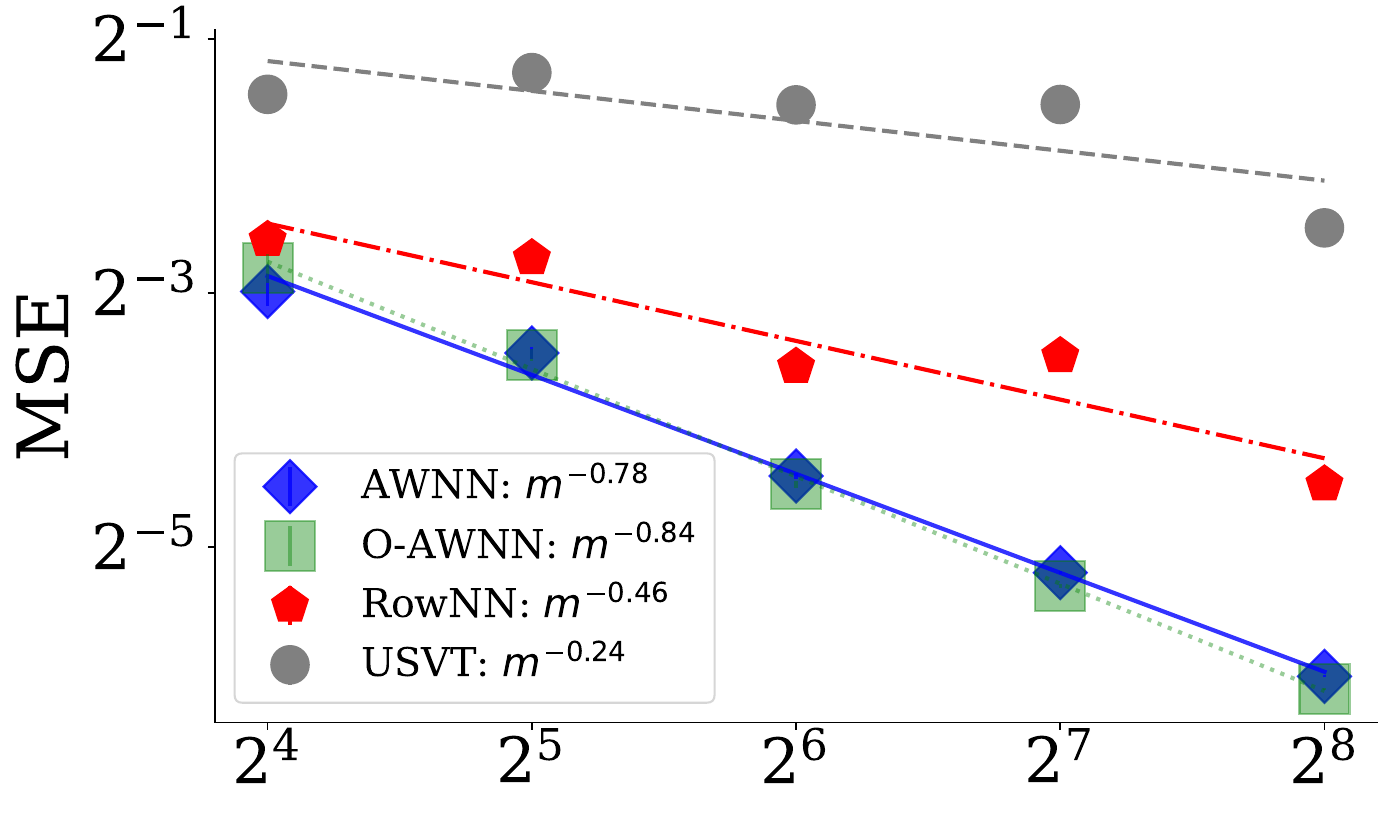} &
        \includegraphics[trim=0in 0in 0in 0in, clip, width=0.47\textwidth]{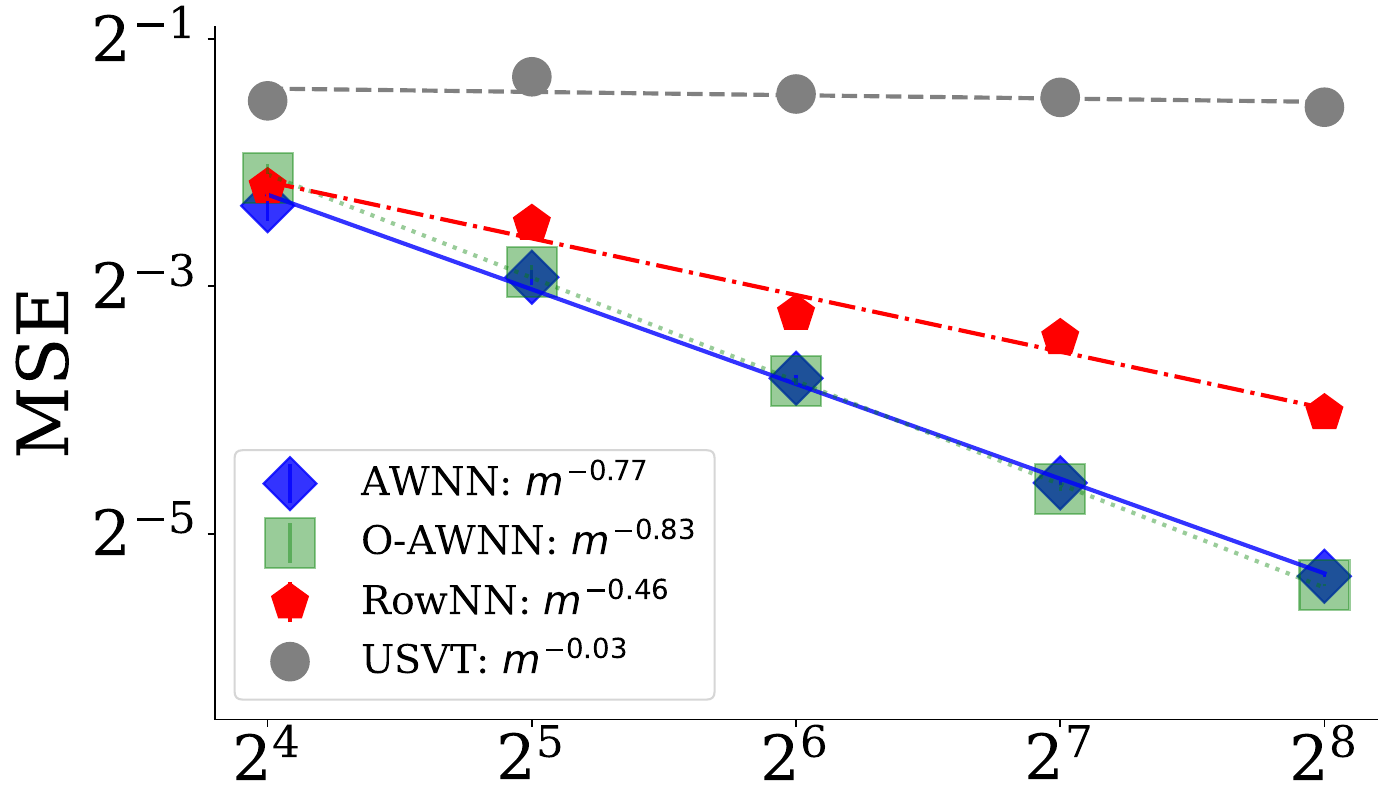}\\[-0.5mm]
        \ \ \quad \# Rows $n$ &\ \  \quad  \# Rows $n$
    \end{tabular}
    \\[2mm]
    
    \caption{\tbf{MSE of $\awnn$ and the benchmarks as a function of number of rows $n(=m)$.} Results are averaged across 10 runs where signals were generated from Lipschitz function ($f$ with $\lambda=1$) with latent variables' dimension $d = 2$ and $\msf{SNR}=1$.}
    \label{fig: mse comparison}
\end{figure*}

\begin{figure*}[ht]
    \centering
    \begin{tabular}{c|cc}
         &\quad\qquad $\msf{SNR} =10$ & \quad \qquad $\msf{SNR} =2$ \\
         \midrule\\
        \raisebox{1.25\height}{\begin{turn}{90}\textbf{MCAR}\end{turn}}  
        &\includegraphics[trim=0in 0in 0in 0in, clip, width=0.47\textwidth]{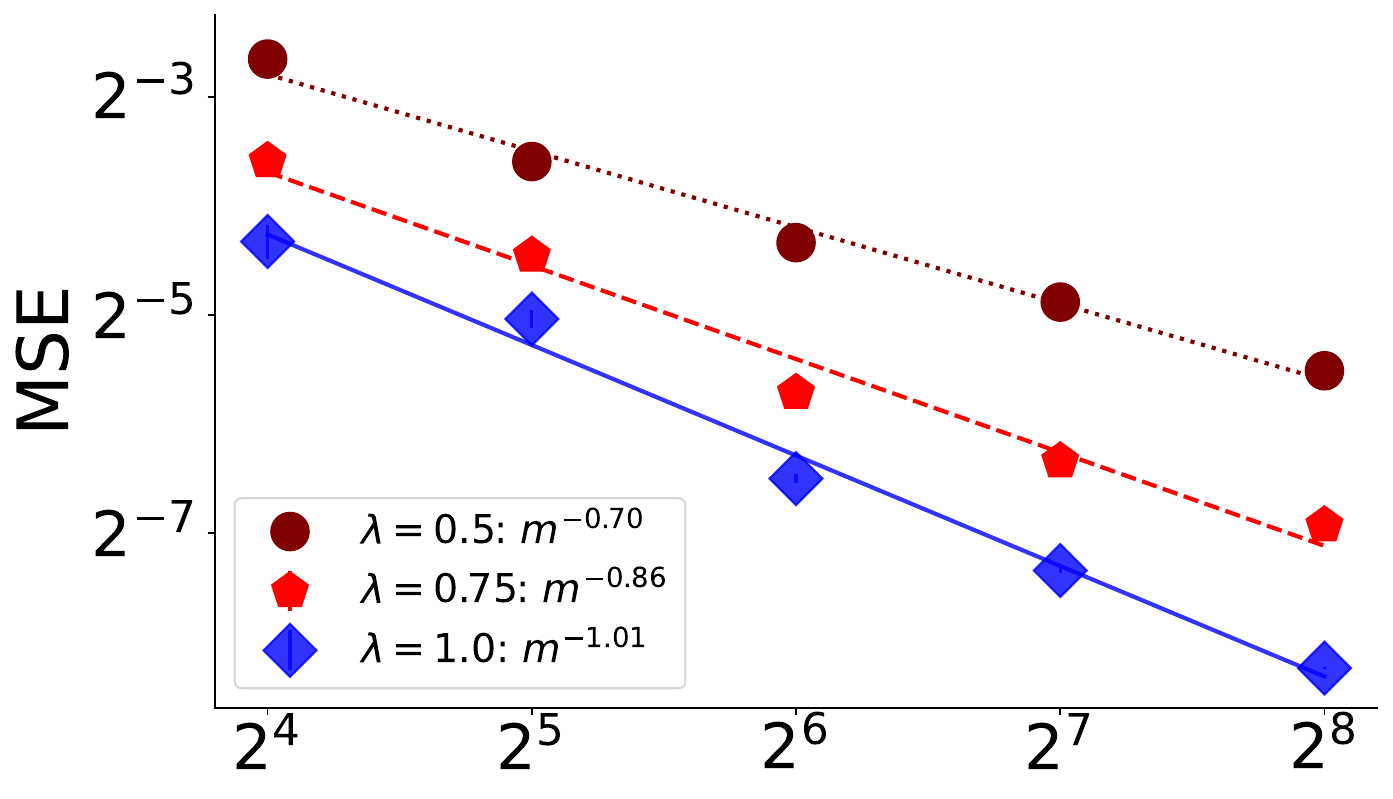} &
        \includegraphics[trim=0in 0in 0in 0in, clip, width=0.47\textwidth]{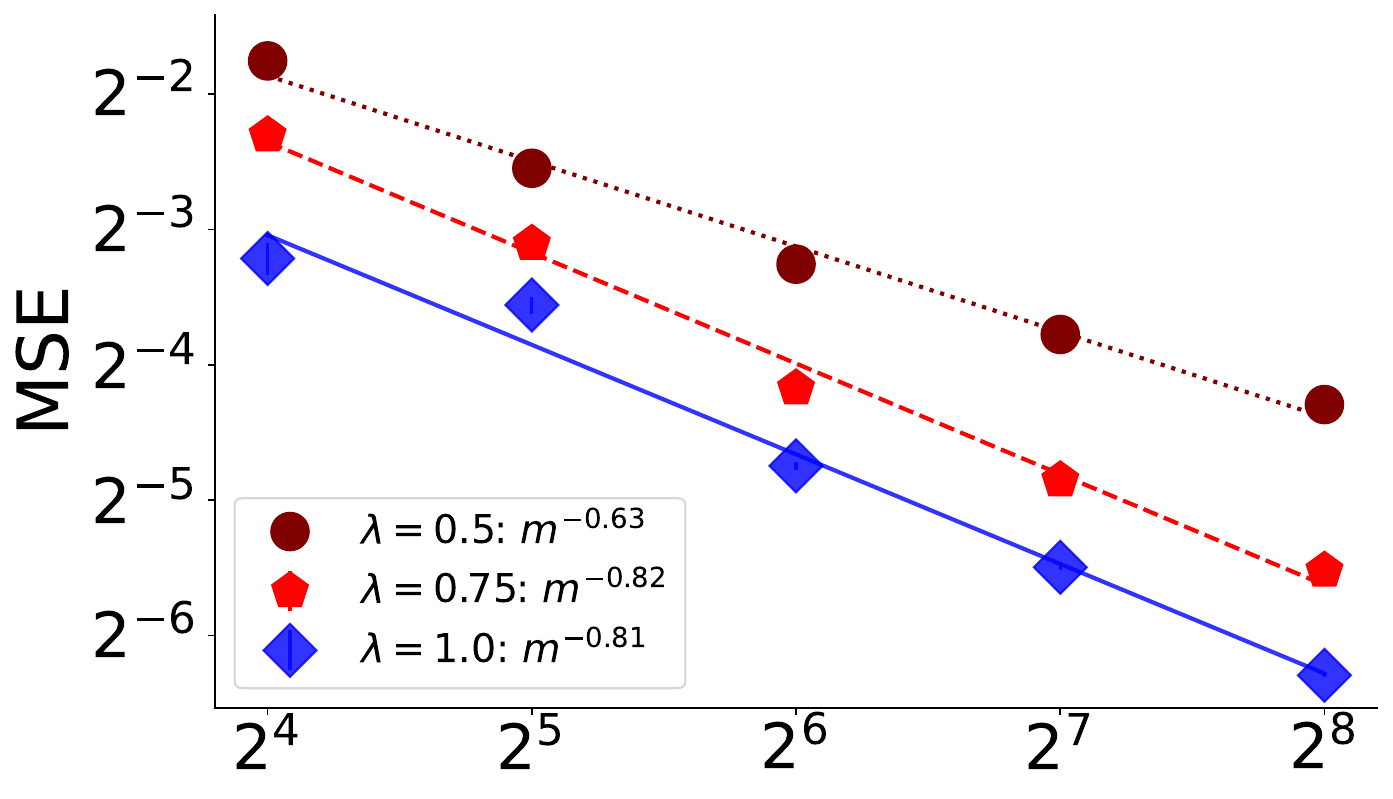}\\[-0.5mm]
        \raisebox{0.2\height}{\begin{turn}{90}\textbf{No Missingness}\end{turn}}
        &\includegraphics[trim=0in 0in 0in 0in, clip, width=0.47\textwidth]{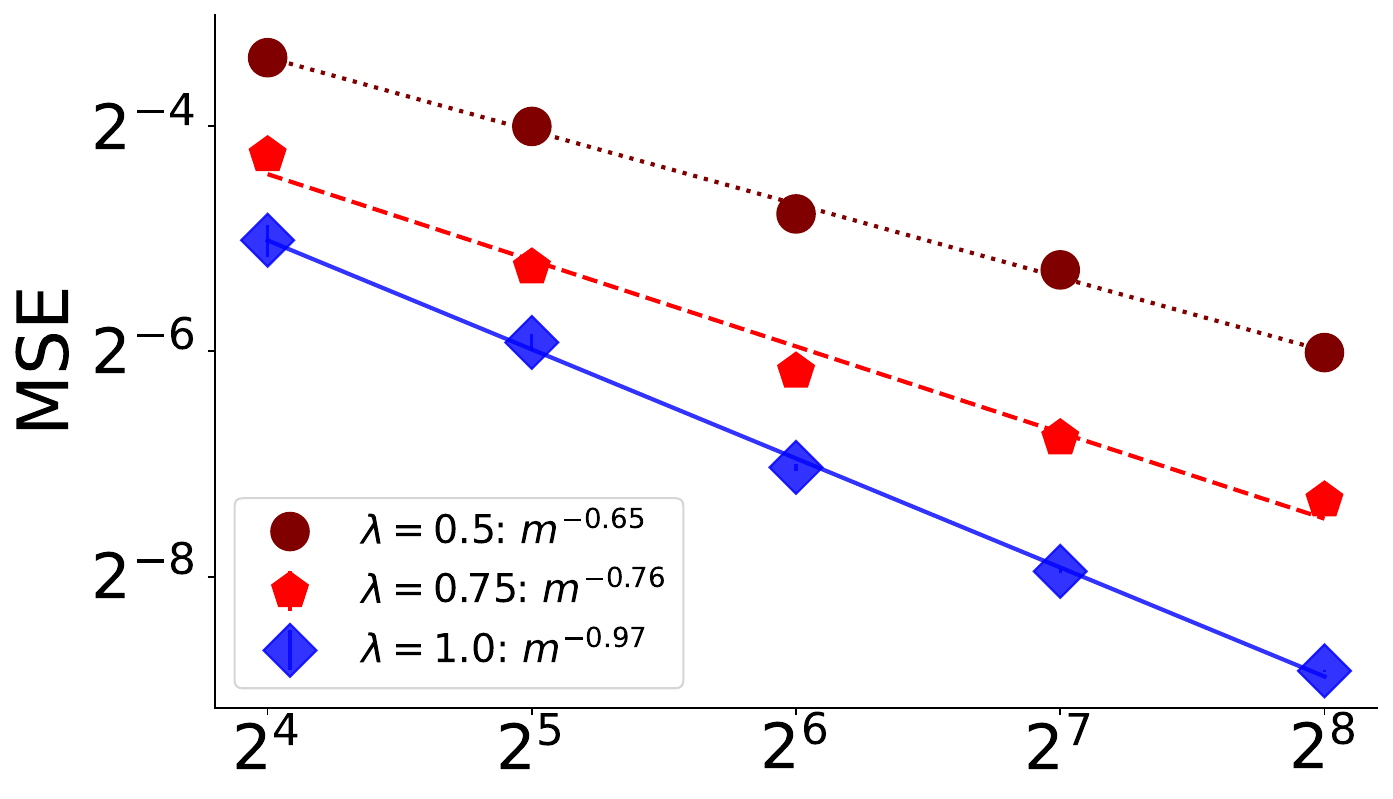} &
        \includegraphics[trim=0in 0in 0in 0in, clip, width=0.47\textwidth]{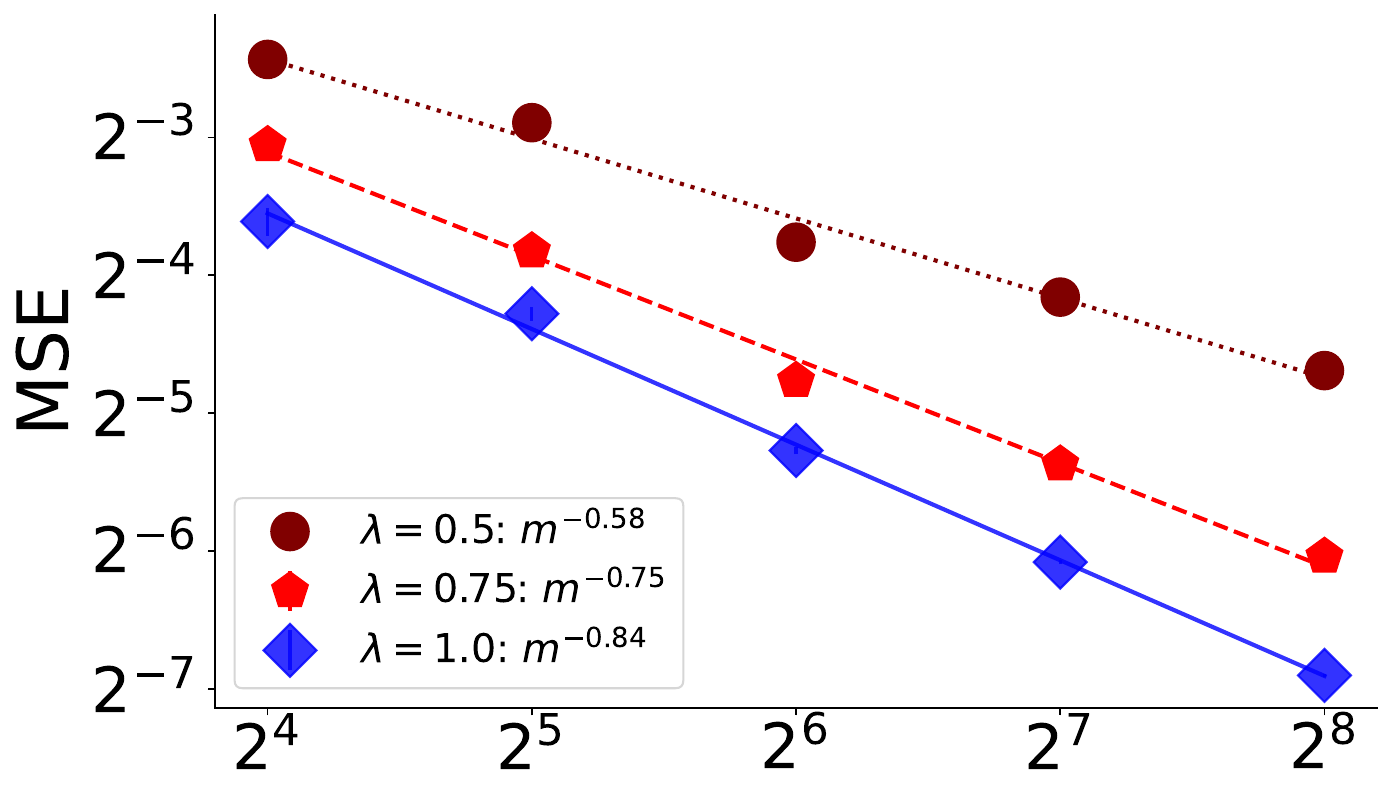}\\[-0.5mm]
        &\ \ \quad \# Rows $n$ &\ \  \quad  \# Rows $n$
    \end{tabular}
    \\[2mm]
    
    \caption{\tbf{Variation of $\awnn$'s MSE behaviour with changing smoothness levels $(\lambda)$ of signals $\braces{\theta_{i,j}}_{i,j\in[n]\times[m]}$ in synthetic data.} Results are averaged across 10 runs where signals were generated from Lipschitz function ($f$ with $\lambda\in\braces{0.5, 0.75, 1}$) with latent variables' dimension $d = 2$. Top and bottom row correspond to MCAR and no missingness setup respectively while left and right column correspond to $\msf{SNR}$ of 10 and 2 respectively.}
    \label{fig: mse vs lambda}
\end{figure*}

\paragraph{Experimental setup} We work with square matrices i.e., $n = m$ to keep the demonstration of empirical results straightforward. RowNN's hyperparameter $\eta$ is tuned via cross-validation with the $\msf{hyperopt}$ Python package. In \cref{fig: mse comparison} and \cref{fig: mse vs lambda}, MSE was recorded for 10 replications and then mean MSE along with 1SD bars were reported. We demonstrate the rate of convergence of the algorithms in synthetic data by calculating the slope of linear regression of $\log\parenth{\mrm{MSE}}$ on $\log(n)$ for each procedure. Note that when slope is -0.5 (say), MSE decay rate of that algorithm is $n^{-0.5}$.

\paragraph{Results} Remarkably, we notice that the MSE decay lines of both $\awnn$ and $\oawnn$ coincide. This indicates the fixed point iteration in \rownnlink estimates the error variance $\sigma_\eps^2$ with pinpoint accuracy. We highlight that \rownnlink achieves this incredible precision on top of being a fully automatic data-driven algorithm, emphasizing empirically the merits of tuning via convex optimization as compared to ad-hoc means like cross validation. Using \cref{cor:gen_missing_MCAR} and ~\cite{dwivedi2022counterfactual}'s Cor. 1(b), we see that both $\awnn$ and RowNN will have theoretical minimax MSE decay rate of $\bigO{n^{-\frac{1}{2}}}$ in \cref{fig: mse comparison} with $\lambda = 1$ and $d=2$. We see empirical validation of \cref{lem:bias_comp} in both the missingness setups of \cref{fig: mse comparison} where even if RowNN nearly achieves its minimax optimal rate, $\awnn$ shows better empirical MSE decay rate. USVT suffers from bigger MSE for no missingness setup and shows trivial MSE decay as soon as MCAR missingness is introduced. Overall, $\awnn$ is superior to the baselines in \cref{fig: mse comparison} and shows the best possible performance as specified by the exact coincidence of $\awnn$ and $\oawnn$'s error graphs.

In \cref{fig: mse vs lambda}, we experimentally verify \cref{cor:exact_mse_rate} and \cref{cor:gen_missing_MCAR} by comparing the error of $\awnn$ for varying levels of smoothness ($\lambda \in\braces{0.5,0.75,1}$) in $f$ \cref{eq: simulation f}. We note that $\awnn$ shows better empirical rates than the corresponding theoretical rates of $\bigO{n^{-\frac{2\lambda}{2\lambda + 2}}}$ in all the setups, also the decay rate improves with increasing smoothness $\lambda$. Our synthetic results are in accordance with ~\cite{sadhukhan2024adaptivity}, as the $\msf{SNR}$ increases, not only does the $\awnn$'s MSE decay rate improves, but also the differences in empirical error rates with changing $\lambda$ become more pronounced.

\bibliographystyle{alpha} 
\bibliography{refs}
\newpage
\appendix
\setcounter{section}{0}
\setcounter{equation}{0}
\setcounter{figure}{0}
\setcounter{remark}{0}
\renewcommand{\thesection}{S.\arabic{section}}
\renewcommand{\theequation}{E.\arabic{equation}}
\renewcommand{\thefigure}{A.\arabic{figure}}
\renewcommand{\theremark}{R.\arabic{remark}}
  \begin{center}
  \Large {\bf Appendix to ``Adaptively Weighted Nearest Neighbors for Matrix Completion''}
  \end{center}

\section{Proof of \Cref{thm:row_wise_bounds}}
\label{app:proof_row_wise_bounds}
Recall our main model \cref{eq:main_model} with all entries observed, i.e., $A_{i,j} = 1$ for all $i,j$. For a given choice of weights $\wgtvector$, the $\awnn$ estimator becomes $\what\ph_{i,j} = \sumn[i']\simplewgt X_{i',j}$ and its row-wise MSE is 
\begin{align}
    \frac{1}{m}\summn[j]\parenth{\simpleth - \what\ph_{i,j}}^2 \stackrel{\cref{eq:main_model}}{=}& \frac{1}{m}\summn[j]\parenth{\simpleth - \sumn[i']\simplewgt (\ph_{i',j} +\eps_{i',j})}^2\\ 
    \stackrel{\label{eq:wgts sum up to one}}{=}&\frac{1}{m}\summn[j]\parenth{\summn[i']\simplewgt (\simpleth - \ph_{i',j} ) - \simplewgt\eps_{i',j}}^2\\
    \leq&\frac{2}{m}\summn[j]\brackets{\parenth{\sumn[i']\simplewgt (\simpleth - \ph_{i',j} )}^2 +\parenth{\summn[i']\simplewgt\eps_{i',j}}^2} \label{eq:prop1 quad ineq}\\
    \leq&\frac{2}{m}\summn[j]\brackets{\underbrace{\sumn[i']\simplewgt (\simpleth - \ph_{i',j} )^2}_{\mbb{B}_{i,j}} +\underbrace{\parenth{\sumn[i']\simplewgt\eps_{i',j}}^2}_{\mbb{V}_{i,j}}}\label{eq:prop1 quad ineq2},
\end{align}

where the step \cref{eq:wgts sum up to one} follow as $\sumn[i']\simplewgt = 1$ and $\simplewgt\in \real_+$. In steps \cref{eq:prop1 quad ineq} and \cref{eq:prop1 quad ineq2}, we use the inequality that for non-negative weights $\{w_i\}_{i =1}^n$ which add up to $1$ and for non-negative quantities $\{x_i\}_{i = 1}^n$, we have $(\sum_{i = 1}^n w_i x_i)^2 \leq \sum_{i = 1}^n w_ix_i^2$. Since $\epsilon_{i,j} \sim \mbox{Sub-Gaussian}(\sigma^2)$ we have,

\begin{align}
    & \sum_{i' = 1}^n \simplewgt\eps_{i',j} \sim \mbox{Sub-Gaussian}(\sigma^2\twonorm{\wgtvector}^2) \\
   \implies & \mathbb{P}\left(\abss{\sum_{i' = 1}^n \simplewgt\eps_{i',j}} > t \right) \leq 2 \exp \left(\frac{-t^2}{2\sigma^2 \twonorm{\wgtvector}^2} \right), \label{eq:prop1 hoeff ineq}
\end{align}

where the step \cref{eq:prop1 hoeff ineq} follows from the Hoeffding Inequality \cite{bentkus2004hoeffding} on the Sub-Gaussian variables. This implies that with probability at-least $1 - 2\delta$, 
\begin{align}
    \label{eq:prop1 var_bound}
   \frac{2}{m}\summn[j]\mathbb{V}_{i,j} =  \frac{2}{m}\summn[j]\left( \sum_{i' = 1}^n \simplewgt \epsilon_{i',j}\right)^2 \leq 4\sigma^2 \twonorm{\wgtvector}^2 \log(2m/\delta).  
\end{align}

For the bias term, we have 
\begin{align}
    \label{eq:prop1 bias_bound}
    \frac{2}{m}\summn[j]\mbb{B}_{i,j} = & \frac{2}{m}\summn[j]\sumn[i']\simplewgt (\simpleth - \ph_{i',j} )^2\\
    = & \frac{2}{m}\sumn[i']\simplewgt \summn[j](\simpleth - \ph_{i',j} )^2\\
    \stackrel{\cref{eq:oracle_row_dist}}{=} & 2\sumn[i']\simplewgt\distroworacle\\
    \stackrel{\cref{assump:dist_concentration}}{\leq} & 2\sumn[i']\simplewgt\parenth{\distrow + \distrowerror} \quad\qtext{with prob. $1-\delta$}\\
    \leq & 2\sumn[i']\simplewgt\distrow[j] + 2\max_{i'\in[n]}\distrowerror\\
\end{align}

Combining all the terms, we have with prob. $1-3\delta$
\begin{align}
    \frac{1}{m}\summn[j]\parenth{\simpleth - \what\ph_{i,j}}^2 \leq 2\braces{2\sigma^2 \twonorm{\wgtvector}^2 \log(2m/\delta) + \sumn[i']\simplewgt\distrow[j]} + 2\max_{i'\in[n]}\distrowerror
\end{align}

\section{Proof of \Cref{thm:optimal_weights}}
\label{app:proof_optimal_weights}

Let $\Delta^+_n$ be the $n$-dimensional probability simplex and $ w^{(i)} = (w_{1}, \cdots, w_{n})$. Then,

\begin{align}
\label{eq:noiseless original optimization prob}
    \widehat w^{(i)}    \coloneqq\arg\min_{w^{(i)}\in \Delta^+_n
    }\brackets{2\log(2m/\delta)\widehat \sigma^2\| w^{(i)}\|_2^2+ \sum_{i' = 1}^n  w_{i'} \widehat \rho_{i', i} }
\end{align}

We transform this problem into an unconstrained convex optimization problem by considering the following program, where $C = 2\log(2m/\delta)$.

\begin{align}
\label{eq: unconstrained optimization prob}
    &\inf_{w^{(i)}
    }\sup_{\lambda\in\real}\brackets{2\log(2m/\delta) \what\sigma^2\| w^{(i)}\|_2^2+ \sum_{i' = 1}^n  \abss{w_{i'}} \widehat \rho_{i', i} - \lambda\parenth{\sumn[i']\simplewgt - 1}}\\
    =&\inf_{w^{(i)}
    }\sup_{\lambda\in\real}\brackets{C\what\sigma^2\| w^{(i)}\|_2^2+ \sum_{i' = 1}^n  \abss{w_{i'}} \widehat \rho_{i', i} - \lambda\parenth{\sumn[i']\simplewgt - 1}}.
\end{align}

 Suppose $(\optwgt,\optlambda)$ is the optimal solution of the \cref{eq: unconstrained optimization prob}. We will show that $\optwgt \in \Delta^+_n$, hence the optimal weights of \cref{eq: unconstrained optimization prob} will also be the optimal solution to the \cref{eq:noiseless original optimization prob}.

From the KKT conditions, differentiating \cref{eq: unconstrained optimization prob} w.r.t. $\lambda$ at $(\optwgt,\optlambda)$ gives:

\begin{align}
\label{eq: opt wgts sum}
    \sumn[i']\optwgt_{i'} = 1 \iff \sum_{j:\optwgt_j \neq 0}\optwgt_{j} = 1
\end{align}

Similarly, differentiating \cref{eq: unconstrained optimization prob} w.r.t. $w_j$ at $(\optwgt,\optlambda)$ gives:

\begin{align}
\label{eq: opt wgt equation}
    &2C\what\sigma^2\optwgtj + \distrow[j] \pderiv{\abss{w_j}}{w_j} - \optlambda = 0\\
    \implies&2C\what\sigma^2\optwgtj + \distrow[j] \biggbraces{\indic{\optwgtj > 0} - \indic{\optwgtj < 0} + v_j\indic{\optwgt_j = 0}} = \optlambda \quad\qtext{(where $v_j\in[-1,1]$)}\\
    \stackrel{\cref{eq: opt wgts sum}}{\implies}& \sumn[j]\indic{\optwgtj\neq 0}\optlambda = 2C\hatsigmasq + \sumn[j]\indic{\optwgtj\neq 0}\distrow[j] \biggbraces{\indic{\optwgtj > 0} - \indic{\optwgtj < 0} + v_j\indic{\optwgt_j = 0}}\\
    \implies & \optlambda = \frac{2C\hatsigmasq}{\sumn[j]\indic{\optwgtj\neq 0}} + \frac{1}{\sumn[j]\indic{\optwgtj\neq 0}}\sumn[j]\distrow[j] \biggbraces{\indic{\optwgtj > 0} - \indic{\optwgtj < 0}}\\
    \iff\quad & \optlambda = \frac{2C\hatsigmasq}{K_i} + \frac{1}{K_i}\sumn[j]\distrow[j] \biggbraces{\indic{\optwgtj > 0} - \indic{\optwgtj < 0}} \quad\qtext{(where $K_i = \sumn[j]\indic{\optwgtj\neq 0}$)}
\end{align}
Rephrasing \cref{eq: opt wgt equation} and substituting $\optlambda$, we have
\begin{align}
    \optwgtj &= \frac{\optlambda}{2C\hatsigmasq} - \frac{\distrow[j] \biggbraces{\indic{\optwgtj > 0} - \indic{\optwgtj < 0} + v_j\indic{\optwgt_j = 0}}}{2C\hatsigmasq} \\
    & = \frac{1}{2C\hatsigmasq}\brackets{\frac{2C\hatsigmasq}{K_i} + \frac{1}{K_i}\sumn[j]\distrow[j] \biggbraces{\indic{\optwgtj > 0} - \indic{\optwgtj < 0}} -\distrow[j] \biggbraces{\indic{\optwgtj > 0} - \indic{\optwgtj < 0} + v_j\indic{\optwgt_j = 0}}}
\end{align}
Suppose $\optwgtj[j']>0$ (such a $j'\in[n]$ will exist otherwise \cref{eq: opt wgts sum} wont be satisfied) then,
\begin{align}
\label{eq: positive wgts formula}
    \optwgtj[j'] = \frac{1}{K_i} - \frac{\distrow[j']}{\deno} + \frac{1}{\deno{K_i}}\sumn[j]\distrow[j] \biggbraces{\indic{\optwgtj > 0} - \indic{\optwgtj < 0}}
\end{align}
Now we will prove via contradiction that $\optwgt\geq 0$. Define 
\begin{align}
    \mathcal{I}_{\mrm{pos}}&\coloneq\braces{k\in [n]:\optwgtj[k]>0}\\
    \mathcal{I}_{\mrm{neg}}&\coloneq\braces{k\in [n]:\optwgtj[k]<0}\\
    \mathcal{I}_{\mrm{nonzero}}&\coloneq\braces{k\in [n]:\optwgtj[k]\neq 0} = \mathcal{I}_{\mrm{pos}} \cup \mathcal{I}_{\mrm{neg}}\\
    \mathcal{I}_{\mrm{zero}}&\coloneq\braces{k\in [n]:\optwgtj[k] = 0}
\end{align}
We note certain facts about the optimal solution to \cref{eq: unconstrained optimization prob},
\begin{align}
    \distrow[j_1]<\distrow[j_2]&\stackrel{\cref{eq: positive wgts formula}}{\iff} \optwgtj[j_1] >\optwgtj[j_2] \quad \forall j_1,j_2\in\mathcal{I}_\mrm{pos}\\
    \forall j_1\in\mathcal{I}_\mrm{pos}\qtext{and} \forall j_2\in\mathcal{I}_\mrm{zero} &\implies \distrow[j_1] \leq\distrow[j_2]\label{eq: pos and zero wgts relation}\\
    \forall j_1\in\mathcal{I}_\mrm{neg}\qtext{and} \forall j_2\in\mathcal{I}_\mrm{zero} &\implies \distrow[j_1] \leq \distrow[j_2]\label{eq: neg and zero wgts relation}
\end{align}
\cref{eq: pos and zero wgts relation} follows as otherwise swapping the $\optwgtj[j_1]$ and  $\optwgtj[j_2]$ in $\optwgt$ will lower the objective function's value in \cref{eq: unconstrained optimization prob}, contradicting the optimality of $\optwgt$. Using a similar argument, we can also prove \cref{eq: neg and zero wgts relation}.

Suppose $\mathcal{I}_{\mrm{neg}}\neq \phi$ i.e., some weights are strictly negative. Choose $k \in \mathcal{I}_\mrm{neg}$, then $\optwgtj[k]$ will be
\begin{align}
    &\optwgtj[k] = \frac{1}{K_i} - \parenth{\frac{-\distrow[k]}{\deno}} + \frac{1}{\deno{K_i}}{\sumn[j]\distrow[j] \biggbraces{\indic{\optwgtj > 0} - \indic{\optwgtj < 0}}}\\
    \stackrel{\cref{eq: positive wgts formula}}{\implies} & \optwgtj[k] - \frac{\distrow[k]}{\deno} = \optwgtj[j'] + \frac{\distrow[j']}{\deno}\qquad\forall j'\in\mathcal{I}_\mrm{pos}\\
    \iff\quad & \optwgtj[k] = \optwgtj[j'] + \frac{\distrow[j']}{\deno} +  \frac{\distrow[k]}{\deno}\qquad\forall j'\in\mathcal{I}_\mrm{pos}\\
    \implies & \optwgtj[k] \geq \optwgtj[j'] > 0\qtext{contradicting the fact that}k\in\mathcal{I}_{\mrm{neg}}
\end{align}
So, $\mathcal{I}_{\mrm{neg}}$ has to be an empty set and $\optwgt \geq 0$. Therefore, by setting $\mathcal{R}(i) = \mathcal{I}_\mrm{pos}$ we have $K_i=\abss{\mathcal{R}(i)}$ and
\begin{align}
     \optwgtj[i'] &= \indic{i'\in\mathcal{R}(i)}\brackets{\frac{1}{K_i} - \frac{\distrow[i']}{\deno} + \frac{1}{\deno{K_i}}\sumn[j]\distrow[j] \indic{\optwgtj > 0} } \quad \mbox{for all} \quad i' \in [n],\\
     &= \indic{i'\in\mathcal{R}(i)}\brackets{\frac{1}{K_i} - \frac{\distrow[i']}{\deno} + \frac{1}{\deno{K_i}}\sum_{j\in\mathcal{R}(i)}\distrow[j] }\\
     &= \indic{i'\in\mathcal{R}(i)}\brackets{\frac{1}{K_i} - \frac{1}{4 \log(2m/\delta)\sigma^2}\parenth{\drown{i}{i'} - \Bar{\widehat \rho}_i}}.
\end{align}





\section{Proof of \Cref{thm:mse_upper_boun_nonmissing}}
\label{app:proof_upper_bound_non_missing}
The row wise mean squared error has the following upper bound (\Cref{thm:row_wise_bounds}) with probability at-least $1 - 3\delta$.
\begin{align}
    \frac{1}{m}\sum_{j = 1}^m (\widehat \theta_{i,j} - \theta_{i,j})^2 \leq 2 \left\{2\log(2m/\delta)\sigma^2\| w^{(i)}\|_2^2 + \sum_{i' = 1}^n  w_{i'} \widehat \rho_{i', i}   \right\} + 2 \max_{i' \in [n]} \zeta_{i',i}(m , \delta). 
\end{align}
Using the representation \Cref{thm:optimal_weights} in the above inequality we can show that the following bound holds with probability at-least $1 - 3\delta$, 
\begin{align}
   \frac{1}{m}\sum_{j = 1}^m (\widehat \theta_{i,j} - \theta_{i,j})^2 \leq& 4 \log(2m/\delta)\sigma^2 \sum_{i' \in \mathcal{R}(i)} \left\{ \frac{1}{K_i^2} - \frac{2}{4 \log(2m/\delta)\sigma^2K_i}( \widehat \rho_{i', i} - \Bar{\widehat \rho}_i) + \frac{1}{(4 \log(2m/\delta)\sigma^2)^2}( \widehat \rho_{i', i} - \Bar{\widehat \rho}_i)^2\right\} \\
   &+ 2\sum_{i' \in \mathcal{R}(i)} \left\{ \frac{\drown{i'}{i}}{K_i} - \frac{\drown{i'}{i}}{4 \log(2m/\delta)\sigma^2}( \widehat \rho_{i', i} - \Bar{\widehat \rho}_i) \right\} + 2 \max_{i' \in [n]} \zeta_{i',i}(m , \delta) \\
   = & 2 \left\{\Bar{\widehat \rho}_i - \frac{\sum_{i' \in \mathcal{R}(i)} \left(\drown{i'}{i} - \Bar{\widehat \rho}_i\right)^2}{8\log(2m/\delta)\sigma^2} + \max_{i' \in [n]} \zeta_{i',i}(m , \delta) \right\}+ \frac{4\log(2m/\delta)\sigma^2}{K_i}.
\end{align}
This completes the proof of the theorem. 

\section{Proof of \Cref{lem:bias_comp}}
\label{app:bias_comp}
Making slight abuse of notation we define weights $\widehat w_{i'}^K$ for all $i' \in [n]$ and for any $K \in [n]$ as, 
\begin{align}
\label{eq:weight_non_missing_general_def}
\widehat w_{i'}^K \coloneqq  \indic{i'\in\mathcal{R}(i)_K}\brackets{\frac{1}{K} - \frac{1}{4 \log(2m/\delta)\sigma^2}\parenth{\drown{i'}{i} - \Bar{\widehat \rho}_i^K}},
\end{align}
where $\mathcal{R}(i)_K$ is the set of $K$ nearest neighboring rows from row $i$ (measured in terms of the emprical distance $\widehat \rho$) and $\Bar{\widehat \rho}_i^K$ (by an abuse of notation) is the mean distance $\widehat \rho$ of the rows in $\mathcal{R}(i)_K$ from row-$i$. It is clear from this definition that for $K = K_i$, the weights $\widehat w_{i'}^{K_i}$ match with the optimal weights $\widehat w_{i'}$ (defined in \Cref{thm:optimal_weights}). Let $K_{max} \in [n]$ be the largest value of $K$ such that the weights $\widehat w_{i'}^K$ lie in $[0,1]$. For any $K \in [0, K_{max}]$ we define MSE$(K)$ as, 
\begin{align}
    \mbox{MSE}(K) = 2 \left\{\Bar{\widehat \rho}_i^K - \frac{\sum_{i' \in \mathcal{R}(i)_K} \left(\drown{i'}{i} - \Bar{\widehat \rho}_i^K\right)^2}{8\log(2m/\delta)\sigma^2} + \max_{i' \in [n]} \zeta_{i',i}(m , \delta) \right\} +  \frac{4\log(2m/\delta)\sigma^2}{K}.
\end{align}
Note that for $K = K_i$, we get back the row-wise mean squared error guarantee for $\awnn$ i.e.\ MSE$(K_i) = \mathbb{B}_i^{\mathrm{AW}} + \mathbb{V}_i^{\mathrm{AW}}$. Moreover following the proofs of \Cref{thm:mse_upper_boun_nonmissing} and \Cref{thm:row_wise_bounds}, one can easily show that the row-wise mean squared error using weighted nearest neighbors with weights $\{\widehat w_{i'}^K\}_{i' = 1}^n$ is bounded above by MSE$(K)$ with probability at-least $1 - 3\delta$. Since $\awnn$ computes the weight vector $\widehat{w}^{(i)}$ by solving the optimization equation \eqref{eq:algo_step_nomissing}, the number of nearest neighbors $K_i$ selected by $\awnn$ satisfies, 
\begin{align}
    K_i  = \mbox{arg max}_{K \in (0, K_{max}]} \mbox{MSE}(K). 
\end{align}
Let $\eta_K$ be the choice of radius of the unweighted row nearest neighbor that leads to the selection of $K $ nearest neighbors i.e.\ $k_{\eta_K^2} = K$. Then for $K \in (0, K_{max}]$ the following inequality holds, 
\begin{align}
\label{eq:mse_aw_user_comp}
    \mbox{MSE}(K_i) \leq & \mbox{MSE}(K) \\
    = & 2 \left\{\Bar{\widehat \rho}_i^K - \frac{\sum_{i' \in \mathcal{R}(i)_K} \left(\drown{i'}{i} - \Bar{\widehat \rho}_i^K\right)^2}{8\log(2m/\delta)\sigma^2} + \max_{i' \in [n]} \zeta_{i',i}(m , \delta) \right\} +  \frac{4\log(2m/\delta)\sigma^2}{K} \\
    \leq & 2 \left\{\Bar{\widehat \rho}_i^K + \max_{i' \in [n]} \zeta_{i',i}(m , \delta) \right\} +  \frac{4\log(2m/\delta)\sigma^2}{K}\\
    = & \mathbb{B}_i^{\mathrm{UW}}(\eta_K^2) + \mathbb{V}_i^{\mathrm{UW}}(\eta_K^2). 
\end{align}
We now show by contradiction that the minimum value of $\mathbb{B}_i^{\mathrm{UW}}(\eta_K^2) + \mathbb{V}_i^{\mathrm{UW}}(\eta_K^2)$ is attained for $K \in (0, K_{max}]$. Suppose the minimum value of $\mathbb{B}_i^{\mathrm{UW}}(\eta_K^2) + \mathbb{V}_i^{\mathrm{UW}}(\eta_K^2)$ is attained at $K > K_{max}$. Since $K > K_{max}$, the weight on the further row in the $K$-nearest neighbor set must be negative. Let $i_K$ denote the $K$-th farthest row from row-$i$ in terms of the empirical distance $\widehat \rho$. We have,
\begin{align}
   &\widehat w_{i_K}^K < 0 \\
\implies & \frac{1}{K} - \frac{1}{4\log(2m/\delta)\sigma^2}(\drown{i_K}{i}  - \Bar{\widehat \rho}_i^K) < 0    \\
\implies & \frac{4\log(2m/\delta)\sigma^2}{K} + \Bar{\widehat \rho}_i^K < \drown{i_K}{i} \\
\implies & \frac{4\log(2m/\delta)\sigma^2}{K} + \frac{(K -1) \Bar{\widehat \rho}_i^{K-1}}{K} < \frac{(K-1)\drown{i_K}{i} }{K} \\
\implies & \drown{i_K}{i}  > \frac{4\log(2m/\delta)\sigma^2}{K-1} + \Bar{\widehat \rho}_i^{K-1} .
\end{align}
This implies the following bound on the row wise MSE of the unweighted row nearest neighbors, 
\begin{align}
   \mathbb{B}_i^{\mathrm{UW}}(\eta_{K}^2) + \mathbb{V}_i^{\mathrm{UW}}(\eta_{K}^2) = & 2 \left\{\Bar{\widehat \rho}_i^K + \max_{i' \in [n]} \zeta_{i',i}(m , \delta) \right\} +  \frac{4\log(2m/\delta)\sigma^2}{K} \\
   = & 2 \left\{\frac{(K-1)\Bar{\widehat \rho}_i^{K-1}}{K} + \frac{ \drown{i_K}{i}}{K}  + \max_{i' \in [n]} \zeta_{i',i}(m , \delta) \right\} +  \frac{4\log(2m/\delta)\sigma^2}{K} \\
   \geq & 2 \left\{\frac{(K-1)\Bar{\widehat \rho}_i^{K-1}}{K} + \frac{ \Bar{\widehat \rho}_i^{K-1}}{K} + \frac{4\log(2m/\delta)\sigma^2}{(K-1)K} + \max_{i' \in [n]} \zeta_{i',i}(m , \delta) \right\} +  \frac{4\log(2m/\delta)\sigma^2}{K} \\
   \geq & 2 \left\{\Bar{\widehat \rho}_i^{K-1} + \max_{i' \in [n]} \zeta_{i',i}(m , \delta)  \right\} + \frac{4(K+1)\log(2m/\delta)\sigma^2 }{K(K-1)} \\
   \geq  & 2 \left\{\Bar{\widehat \rho}_i^{K-1} + \max_{i' \in [n]} \zeta_{i',i}(m , \delta)  \right\} +  \frac{4\log(2m/\delta)\sigma^2}{K-1} \\
   = &  \mathbb{B}_i^{\mathrm{UW}}(\eta_{K-1}^2) + \mathbb{V}_i^{\mathrm{UW}}(\eta_{K-1}^2). 
\end{align}
This clearly contradicts the statement that the minimum value of $\mathbb{B}_i^{\mathrm{UW}}(\eta_K^2) + \mathbb{V}_i^{\mathrm{UW}}(\eta_K^2)$ is attained at $K > K_{max}$. Consequently the minimum value of $\mathbb{B}_i^{\mathrm{UW}}(\eta_K^2) + \mathbb{V}_i^{\mathrm{UW}}(\eta_K^2)$ is attained at some $K \in (0, K_{max}]$. Combining this result with \eqref{eq:mse_aw_user_comp} we obtain, 
\begin{align}
    \mathbb{B}_i^{\mathrm{AW}} + \mathbb{V}_i^{\mathrm{AW}} = &  \mbox{MSE}(K_i) \\
    \leq & \min_{K \in [n]}  \{\mathbb{B}_i^{\mathrm{UW}}(\eta_K^2) + \mathbb{V}_i^{\mathrm{UW}}(\eta_K^2)\} \\
    = & \min_{\eta > 0} \{\mathbb{B}_i^{\mathrm{UW}}(\eta^2) + \mathbb{V}_i^{\mathrm{UW}}(\eta^2)\}.
\end{align}
This completes the proof of the lemma. 
\section{Proof of \Cref{cor:exact_mse_rate}}
\label{app:exact_mse_rate_non_missing}
We observe from \Cref{lem:bias_comp} that the following bound holds with probability at-least $1 -3\delta$, 
\begin{align}
\frac{1}{m}\sum_{j = 1}^m (\widehat \theta_{i,j} - \theta_{i,j})^2 \leq &\min_{\eta >0}\left\{ \mathbb{B}_i^{\mathrm{UW}}(\eta^2) + \mathbb{V}_i^{\mathrm{UW}}(\eta^2)\right\} \\
   = &\min_{\eta >0}\left\{  2 \left\{\frac{\sum_{i' \in \neighbors(\eta^2)}\drown{i'}{i}}{k_{\eta^2}} + \max_{i' \in [n]} \zeta_{i',i}(m , \delta) \right\}+  \frac{4\log(2m/\delta)\sigma^2}{k_{\eta^2}} \right\} \\
   \leq & \min_{\eta >0}\left\{ 2 \eta^2 + \frac{4\log(2m/\delta)\sigma^2}{k_{\eta^2}} + 2 \max_{i' \in [n]} \zeta_{i',i}(m , \delta) \right\}. 
\end{align}
We use Lemma-$2$ in \cite{sadhukhan2024adaptivity} to provide an appropriate lower bound on the number of nearest neighbors $k_{\eta^2}$ selected by the unweighted row nearest neighbors with radius parameter $\eta$. It can be shown that under assumptions~\ref{assump:dist_concentration}, \ref{assump:sub_g_noise}, \ref{assump:low_dimensional}, and \ref{assump:row_latent} with probability at-least $1 - \delta$, 
\begin{align}
    k_{\eta^2} \geq (1 - \delta)n \left(\frac{\eta^2- \max_{i' \in [n]} \zeta_{i',i}(m , \delta)}{L^2} \right)^{\frac{d_1}{2\lambda}}. 
\end{align}
Therefore with probability at-least $1 - 4\delta$ the row-wise mean squared error of $\awnn$ satisfies,
\begin{align}
  \frac{1}{m}\sum_{j = 1}^m (\widehat \theta_{i,j} - \theta_{i,j})^2 \leq \min_{\eta >0}\left\{ 2 \eta^2 + \frac{4\log(2m/\delta)\sigma^2}{(1 - \delta)n \left(\frac{\eta^2- \max_{i' \in [n]} \zeta_{i',i}(m , \delta)}{L^2} \right)^{\frac{d_1}{2\lambda}}} + 2 \max_{i' \in [n]} \zeta_{i',i}(m , \delta) \right\}.  
\end{align}
From the expression of row-wise MSE it can be seen that under the regime $n = O(\{ \max_{i' \in [n]} \zeta_{i',i}(m , \delta)\}^{-\frac{d_1 + 2\lambda}{2 \lambda}})$ (this ensures $\eta^2 \gtrsim \max_{i' \in [n]} \zeta_{i',i}(m , \delta)$), the radius $\eta = O(n^{-\lambda/(d_1 + 2\lambda)})$ achieves the minimum value in the row-wise MSE upper bound, 
\begin{align}
    \frac{1}{m}\sum_{j = 1}^m (\widehat \theta_{i,j} - \theta_{i,j})^2  = O\left(n^{-2\lambda/(d_1 + 2\lambda)} \right) . 
\end{align}
Under the other regime since $\max_{i' \in [n]} \zeta_{i',i}(m , \delta) \gtrsim \eta^2 $ we get the trivial rate, 
\begin{align}
    \frac{1}{m}\sum_{j = 1}^m (\widehat \theta_{i,j} - \theta_{i,j})^2  = O\left(\max_{i' \in [n]} \zeta_{i',i}(m , \delta) \right).
\end{align}
This completes the proof of the corollary. 

\section{Proof of \Cref{lem:distance_bound}}
\label{app:distance_bound_awnn}
We simplify the notation by setting $c = 4\log(2m\delta)$ in this proof. From the definition of the weights we have the following for all $i' \in [n]$, 
\begin{align}
\label{eq:dist_emp_ineq}
   & 0 \leq  \what{w}_{i',j}  \leq 1\\
\implies &  \drownmean{j} - \left(1 - \frac{1}{\K[j]} \right)c\sigma^2 \leq  \drown{i'}{i} \leq   \drownmean{j} + \frac{c\sigma^2 }{\K[j]}.
\end{align}
Let $\drown{i_{(1)}}{i}, \cdots, \drown{i_{(\K[j])}}{i}$ be all the empirical row distances $\{\drown{i'}{i}\}_{i' \in \NR[j]}$ arranged in ascending order. If $\K[j] = 2$ the inequality \eqref{eq:dist_emp_ineq} yields, 
\begin{align}
    & \drownmean{j} - \left(1 - \frac{1}{\K[j]} \right)c\sigma^2 \leq \drown{i_{(1)}}{i}  \\
    \implies & \drownmean{j} \leq \frac{c \sigma^2}{2} + \drown{i_{(1)}}{i} \\
    \implies & \frac{\drown{i_{(1)}}{i} + \drown{i_{(2)}}{i}}{2} \leq \frac{c \sigma^2}{2} + \drown{i_{(1)}}{i} \\
    \implies & \drown{i_{(2)}}{i} \leq c \sigma^2 + \drown{i_{(1)}}{i}.
\end{align}
Therefore the first result of \Cref{lem:distance_bound} holds if $\K[j] = 2$. For general $\K[j]$ we prove the result by induction. Suppose $\drow[i_{(k)}] \leq c \sigma^2 + \drow[i_{(1)}]$ for $2 \leq k < \K[j]$. Then we can bound $\drown{i_{(\K[j])}}{i}$ using \eqref{eq:dist_emp_ineq}. 
\begin{align}
   & \drown{i_{(\K[j])}}{i} \leq \drownmean{j} + \frac{c\sigma^2 }{\K[j]} \\
  \implies &   \drown{i_{(\K[j])}}{i} \leq \frac{1}{\K[j] - 1} \sum_{k < \K[j]} \drown{i_{(k)}}{i}  + \frac{c\sigma^2 }{\K[j] - 1} \\
   \implies &   \drown{i_{(\K[j])}}{i}\leq \frac{1}{\K[j] - 1} \left((\K[j] - 2)c\sigma^2 + (\K[j] - 1)\drown{i_{(1)}}{i} \right)  + \frac{c\sigma^2 }{\K[j] - 1} \\
   \implies &   \drown{i_{(\K[j])}}{i}\leq c \sigma^2 + \drown{i_{(1)}}{i}. 
\end{align}
The proof of the first result of \Cref{lem:distance_bound} is completed by noting that for $A_{i, j} = 1$,  $\drown{i_{(1)}}{i} = \min_{i' \in \NR[j]}\drown{i'}{i} = 0$. The second result stated in the lemma is again just a consequence of the definition in \eqref{eq: weight defn}. Rearranging the terms in \eqref{eq: weight defn} we have the following for all $i' \in \NR[j]$,
\[
\drownmean{j} = c\sigma^2 \left( \what{w}_{i',j} - \frac{1}{\K[j]}\right) + \drown{i'}{i}.
\]
In particular this equality holds true for $i' = i_{(1)}$ and we know that $\drown{i_{(1)}}{i} = \min_{i' \in \NR[j]} \drown{i'}{i}$ and $\what{w}_{i_{(1)},j} = \infnorm{\what{w}^{(i, j)}}$. This completes the proof of the lemma.

\section{Proof of \Cref{thm:mse_general_bound}}
\label{app:mse_general_bound}
To aid the demonstration of the upper bound on the row wise MSE in the presence of missingness, we define the set $\mathcal{C}(i')$ for each row $i' \in [n]$, 
\begin{align}
    \NC[i'] =\braces{j \in [m]: i' \in \NR[j]}.
\end{align}
The set $\NC[i']$ consists exactly of those columns $j$ whose neighborhood $\mathcal{R}(i, j)$ contains $i'$. We begin the proof by stating the following lemma. 
\begin{lemma}
\label{lem: quadratic equality}
    For 2 vectors $a$ and $b$, we have the following identity $\forall \lambda\in [0,1]$
    \begin{align}
        \abss{\lambda a+\parenth{1-\lambda}b}^2 = \lambda \abss{a}^2 + \parenth{1-\lambda}\abss{b}^2 - \lambda\parenth{1-\lambda}\abss{a-b}^2
    \end{align}
\end{lemma}

\Cref{lem: quadratic equality} can be easily proved by expanding the square. We obtain the following bias-variance decomposition of the pointwise error, 
\begin{align}
    &(\widehat \theta_{i,j} - \theta_{i,j})^2 \\
= & \left(\sum_{i' = 1}^n  \widehat w_{i',j}A_{i',j} \theta_{i', j} + \sum_{i' = 1}^n  \widehat w_{i',j}A_{i',j} \eps_{i', j}  - \theta_{i,j}\right)^2 \\
\stackrel{(i)}{=} & \left(\sum_{i' = 1}^n  \widehat w_{i',j}\theta_{i', j} + \sum_{i' = 1}^n  \widehat w_{i',j}\eps_{i', j}  - \theta_{i,j}\right)^2 \\
\leq &  2\left(\sum_{i = 1}^n \widehat w_{i,j}^{(i_0)} (\theta_{i, j} - \theta_{i_0,j} ) \right)^2 + 2 \left( \sum_{i' = 1}^n \widehat w_{i',j} \epsilon_{i',j}\right)^2   \\
\stackrel{(ii)}{=} &  2\sum_{i' = 1}^n \widehat w_{i',j}(\theta_{i', j} - \theta_{i,j} )^2 -  2\sum_{i' = 1}^n \sum_{i' < i'' \leq n } \widehat w_{i',j}\widehat w_{i'',j}  (\theta_{i', j}  - \theta_{i'', j} )^2 +  2\left( \sum_{i' = 1}^n \widehat w_{i',j} \epsilon_{i',j}\right)^2 \\
= &2(\mathbb{B}_1)_{i,j}-2(\mathbb{B}_2)_{i,j}+2\mathbb{V}_{i,j}.
\end{align}
Here the step-$(i)$ follows from the fact that $\widehat w_{i',j} A_{i',j} = \widehat w_{i',j}$ for all $i' \in [n]$. In step-$(ii)$ we use the property that for non-negative weights $\{w_i\}_{i =1}^n$ which add up to $1$ and for non-negative integers $\{x_i\}_{i = 1}^n$, we have $(\sum_{i = 1}^n w_i x_i)^2 = \sum_{i = 1}^n w_ix_i^2 - \sum_{1 \leq i < j \leq n} w_iw_j(x_i - x_j)^2$ from \cref{lem: quadratic equality}. Since $\epsilon_{i',j} \sim \mbox{Sub-Gaussian}(\sigma^2)$ we have, 
\begin{align}
    & \sum_{i' = 1}^n \widehat w_{i',j}\epsilon_{i',j} \sim \mbox{Sub-Gaussian}(\sigma^2\sum_{i' = 1}^n  (\widehat w_{i',j})^2) \\
   \implies & \mathbb{P}\left(|\sum_{i' = 1}^n \widehat w_{i',j} \epsilon_{i',j} | > t  \right) \leq 2 \exp \left(\frac{-t^2}{2\sigma^2 \twonorm{w^{(i, j)}}^2} \right),
\end{align}
This implies that with probability at-least $1 - (\delta/m)$ the following event holds, 
\begin{align}
    \label{eq:var_bound}
   \mathbb{V}_{i,j} =  \left( \sum_{i = 1}^n \widehat w_{i,j}^{(i_0)} \epsilon_{i,j}\right)^2 \leq 2\sigma^2 \twonorm{w^{(i_0)}_{\cdot,j}}^2 \log(2m/\delta) = C_{\delta}\sigma^2 \twonorm{w^{(i_0)}_{\cdot,j}}^2 ,  
\end{align}
where $C_{\delta} = 2 \log(2m/\delta)$. Using the representation in \cref{eq: weight defn}, we can bound the variance part as,
\begin{align}
    \mathbb{V}_{i,j} \leq &C_{\delta}\sigma^2 \twonorm{w^{(i, j)}}^2 \\
    = & \frac{C_{\delta}\sigma^2}{\K[j]}+\frac{\sum_{i'\in \NR[j]}\parenth{\drown{i'}{i} - \drownmean{j}}^2}{2C_{\delta}\sigma^2}\\
    = & C_{\delta}\sigma^2 \left\{ \frac{1}{\K[j]}+\sum_{i' \in \NR[j]}\frac{\drown{i'}{i}}{C_{\delta}\sigma^2}\parenth{\frac{\drown{i'}{i} }{2C_{\delta}\sigma^2}-\frac{\drownmean{j}}{2C_{\delta}\sigma^2}} \right\}
\end{align}
Aggregating the variance terms across the row $i$, we observe that the following event holds with probability at-least $1 - \delta$,
\begin{align}
    \label{eq:var_mean_expanse}
    &\frac{2}{m}\sum_{j=1}^m\mathbb{V}_{i,j} \\
    \leq & \frac{2}{m}\sum_{j=1}^m\parenth{\frac{C_{\delta}\sigma^2}{\K[j]}+\sum_{i'\in \NR[j]}\drown{i'}{i}\parenth{\frac{\drown{i'}{i} }{2C_{\delta}\sigma^2}-\frac{\drownmean{j}}{2C_{\delta}\sigma^2}}}\\
    = & \frac{2}{m}\sum_{j=1}^m\frac{C_{\delta}\sigma^2}{\K[j]}+\sum_{i' = 1}^n\frac{2\drown{i'}{i}}{m}\brackets{\sum_{j\in \NC}\frac{1}{\K}} -\sum_{i' = 1}^n\frac{2\drown{i'}{i}}{m}\brackets{\sum_{j\in \NC}\widehat w_{i', j}}\\
\end{align}


We now shift our attention to the first bias term $(\mathbb{B}_1)_{i,j}$ in the decomposition of the pointwise error. We observe the following regarding the bias term $(\mathbb{B}_1)_{i,j}$ averaged over all the columns $j \in [m]$ (for the row $i$),
\begin{align}
\label{eq:bias_decomp_init}
    \frac{1}{m}\sum_{j=1}^m(\mathbb{B}_1)_{i, j}  = & \frac{1}{m}\sum_{j=1}^m \sum_{i' = 1}^n \widehat w_{i',j} (\theta_{i', j} - \theta_{i,j} )^2 \\
    = & \frac{1}{m}\sum_{j=1}^m \sum_{i' \in \NR[j]} \widehat w_{i',j} (\theta_{i', j} - \theta_{i,j} )^2\\
    = & \frac{1}{m} \sum_{i' = 1}^n \sum_{j \in \NC[i']} \widehat w_{i',j}(\theta_{i', j} - \theta_{i,j} )^2 \\
    = & \frac{1}{m} \sum_{i' = 1}^n (\sum_{j \in \NC[i']} \widehat w_{i',j}) \left(\frac{\sum_{j \in \NC[i']} \widehat w_{i',j}(\theta_{i', j} - \theta_{i,j} )^2}{\sum_{j \in \NC[i']} \widehat w_{i',j}} -  \rho_{i',i} \right) + \frac{1}{m} \sum_{i' = 1}^n (\sum_{j \in \NC[i']} \widehat w_{i',j})  \rho_{i',i} \\
    = & \frac{1}{m} \sum_{i' = 1}^n (\sum_{j \in \NC[i']} \widehat w_{i',j}) \left(\tdrown{i'}{i} -  \rho_{i',i} \right) + \frac{1}{m} \sum_{i' = 1}^n (\sum_{j \in \NC[i']} \widehat w_{i',j})  \rho_{i',i},
\end{align}
where, 
\begin{align}
    \tdrown{i'}{i} \coloneqq \frac{\sum_{j \in \NC[i']} \widehat w_{i',j}(\theta_{i', j} - \theta_{i,j} )^2}{\sum_{j \in \NC[i']} \widehat w_{i',j}} \quad \mbox{for all} \quad i' \in [n]. 
\end{align}
For the purpose of proof let us define the population mean, 
\begin{align}
    \rho^*_{i',i} = \mathbb{E}\left[(\theta_{i,j} - \theta_{i', j})^2 \right].
\end{align} 
We also define the stopping times $t_{(l)}(i', i)$ indicating the presence of entries in both rows $i', i$ for the $l$-th time, 
\begin{align}
    t_{(l)}(i', i) = \begin{cases}
        \min\{t: t_{(l-1)}(i', i) < t \leq m \mbox{ such that } t \in \NC[i']\} \quad &\mbox{if such a } t  \mbox{ exists},\\
        m+1 & \mbox{otherwise}. 
    \end{cases}
\end{align}
We observe that $\tdrown{i'}{i} -  \rho^*_{i',i} $ has the following representation, 
\begin{align}
    \tdrown{i'}{i} -  \rho^*_{i',i} = & \frac{\sum_{j \in \NC[i']} \widehat w_{i',j}(\theta_{i', j} - \theta_{i,j} )^2}{\sum_{j \in \NC[i']} \widehat w_{i',j}} -  \rho^*_{i',i} \\
    = & \frac{\sum_{j \in \NC[i']} \widehat w_{i',j}\left[(\theta_{i', j} - \theta_{i,j} )^2 -  \rho^*_{i',i}\right]}{\sum_{j \in \NC[i']} \widehat w_{i',j}} \\
    = & \frac{\sum_{l = 1}^{T_{i', i}} \widehat w_{i',t_{(l)}(i', i)}\indic{t_{(l)}(i', i) \leq m}\left[(\theta_{i', t_{(l)}(i', i)} - \theta_{i,t_{(l)}(i', i)} )^2 -  \rho^*_{i',i}\right]}{\sum_{j \in \NC[i']} \widehat w_{i',j}}\\
    = & \frac{\sum_{l = 1}^{T_{i', i}}W_l }{\sum_{j \in \NC[i']} \widehat w_{i',j}},
\end{align}
where $T_{i',i} = |\NC[i']|$ and $W_l =  \widehat w_{i',t_{(l)}(i', i)} \indic{t_{(l)}(i', i) \leq m}[(\theta_{i', t_{(l)}(i', i)} - \theta_{i,t_{(l)}(i', i)} )^2 -  \rho^*_{i',i}]$ for $l = 1, \cdots,T_{i',i}  $. Since $|\theta_{i,j}| \leq M$ for $i \in [n], j \in [m]$, $|W_l|$ is bounded above by $8M^2  \widehat w_{i',t_{(l)}(i', i)}$ for $l \in [T_{i',i}]$. Let us denote the sigma algebra containing all the information upto time $t$ as $\mathcal{F}_t$ for $t \in [m]$. We denote the sigma field generated by the stopping time $t_{(l)}(i', i)$ by $\mathcal{H}_l$ for $l \in [T_{i',i}]$. We observe that, 
\begin{align}
    \mathbb{E}[W_l| \mathcal{H}_l] = & \mathbb{E}[ \widehat w_{i',t_{(l)}(i', i)}\indic{t_{(l)}(i', i) \leq m}[(\theta_{i', t_{(l)}(i', i)} - \theta_{i,t_{(l)}(i', i)} )^2 -  \rho^*_{i',i}]| \mathcal{H}_l ] \\
    = & \indic{t_{(l)}(i', i) \leq m}\mathbb{E}[ \widehat w_{i',t_{(l)}(i', i)}| \mathcal{H}_l] \mathbb{E}[[(\theta_{i', t_{(l)}(i', i)} - \theta_{i,t_{(l)}(i', i)} )^2 -  \rho^*_{i',i}]| \mathcal{H}_l ] \\
    = &  \indic{t_{(l)}(i', i) \leq m} \mathbb{E}[ \widehat w_{i',t_{(l)}(i', i)}| \mathcal{H}_l](\rho^*_{i',i} - \rho^*_{i',i}) \\
    = & 0. 
\end{align}
Therefore we conclude that $\{W_l\}_{l= 0}^{T_{i',i}}$ is a bounded martingale difference sequence with respect to the sigma algebra $\{\mathcal{H}_l\}_{l= 0}^{T_{i',i}}$. We use the weighted Azuma martingale concentration bound to show that $\tdrown{i'}{i}$ concentrates around $\rho^*_{i', i}$. 
\begin{result}[Weighted Azuma martingale concentration]
\label{result_azuma}
Consider a bounded martingale difference sequence $\{ S_n \}_{n=1}^{\infty}$ adapted to the filtration $\{ \mathcal{F}_n \}_{n = 1}^{\infty}$ i.e.\ $\mathbb{E}[S_n | \mathcal{F}_{n - 1}] = 0 $ for all $n \in \mathbb{N}$. Suppose $|S_n| \leq M_n$ for all $n \in \mathbb{N}$. Then the following event holds with probability at least $ 1- \delta$, 
\begin{align}
    \abss{\sumn S_i} \leq  \sqrt{ \log(2/\delta) \sum_{i = 1}^n M_i^2}. 
\end{align}
\end{result}
Using \Cref{result_azuma} on the bounded martingale difference sequence $\{W_l\}_{l= 0}^{T_{i',i}}$ guarantees that the following event holds with probability at least $1 - \delta$, 
\begin{align}
    \left|\sum_{l = 1}^{T_{i', i}}W_l\right| \leq \sqrt{\log(2/\delta) \sum_{l = 1}^{^{T_{i', i}}} 64M^4 \widehat w_{i',t_{(l)}(i', i)}^2} = 8M^2\sqrt{\log(2/\delta)}\|\widehat w_{i',\cdot}\|_2. 
\end{align}
Therefore the following concentration holds with probability at-least $1 - \delta$, 
\begin{align}
    \left|\tdrown{i'}{i} -  \rho^*_{i',i} \right| = &  \frac{\left| \sum_{l = 1}^{T_{i', i}}W_l  \right|}{\sum_{j \in \NC[i']} \widehat w_{i',j}}    \\
    \leq & 8M^2 \sqrt{\log(2/\delta)} (\|\widehat w_{i',\cdot}\|_2/ \|\widehat w_{i',\cdot}\|_1). 
\end{align}
Following the same steps we can show that with probability at-least $1- \delta$, 
\begin{align}
    \left|\rho_{i', i} -  \rho^*_{i',i} \right| 
    \leq  8M^2 \sqrt{\log(2/\delta)} (\|\widehat w_{i',\cdot}\|_2/ \|\widehat w_{i',\cdot}\|_1). 
\end{align}
Combining the two concentration bounds above we see that the following upper bound holds with probability at-least $1- 2\delta$,
\begin{align}
    \left|\tdrown{i'}{i} - \rho_{i', i}  \right| 
    \leq  16M^2 \sqrt{\log(2/\delta)} (\|\widehat w_{i',\cdot}\|_2/ \|\widehat w_{i',\cdot}\|_1). 
\end{align}
We use this concentration of $\tdrown{i'}{i}$ about $\rho_{i', i}$ to bound the row wise average bias with probability at-least $1 - 2\delta$, 
\begin{align}
    \label{eq:bias_1_mean_expanse}
    \frac{1}{m}\sum_{j=1}^m(\mathbb{B}_1)_{i, j}  
    = & \frac{1}{m} \sum_{i' = 1}^n (\sum_{j \in \NC[i']} \widehat w_{i',j}) \left(\tdrown{i'}{i} -  \rho_{i',i} \right) + \frac{1}{m} \sum_{i' = 1}^n (\sum_{j \in \NC[i']} \widehat w_{i',j})  \rho_{i',i} \\
    \leq & \frac{1}{m} \sum_{i' = 1}^n (\sum_{j \in \NC[i']} \widehat w_{i',j}) (16M^2 \sqrt{\log(2/\delta)} (\|\widehat w_{i',\cdot}\|_2/ \|\widehat w_{i',\cdot}\|_1)) + \frac{1}{m} \sum_{i' = 1}^n (\sum_{j \in \NC[i']} \widehat w_{i',j})  \rho_{i',i} \\
    = & \frac{1}{m} \sum_{i' = 1}^n \left(16M^2 \sqrt{\log(2/\delta)} (\|\widehat w_{i',\cdot}\|_2/ \|\widehat w_{i',\cdot}\|_1) + \rho_{i',i} \right)(\sum_{j \in \NC[i']} \widehat w_{i',j}). 
\end{align}
We put together all the prior analysis to show that the row wise mean squared error of $\awnn$ has the following data-dependent upper boound with probability at-least $1 - 4\delta$, 
\begin{align}
    &\frac{1}{m}\sum_{j=1}^m(\widehat \theta_{i,j} - \theta_{i,j})^2 \\
    \stackrel{(i)}{\leq} & \frac{1}{m}\left(\sum_{j=1}^m2(\mathbb{B}_1)_{i,j}+2 \mathbb{V}_{i,j} \right)\\
    \stackrel{\eqref{eq:bias_1_mean_expanse}}{\leq} &\frac{2}{m} \sum_{i' = 1}^n (16M^2 \sqrt{\log(2/\delta)} (\|\widehat w_{i',\cdot}\|_2/ \|\widehat w_{i',\cdot}\|_1))(\sum_{j \in \NC[i']} \widehat w_{i',j})+ \frac{2}{m}\sum_{i' = 1}^n \rho_{i', i} (\sum_{j \in \NC[i']} \widehat w_{i',j}) +\frac{2}{m}\sum_{j = 1}^m \mathbb{V}_{i,j} \\
    \stackrel{\eqref{eq:var_mean_expanse}}{\leq} & \frac{2}{m} \sum_{i' = 1}^n (16M^2 \sqrt{\log(2/\delta)} (\|\widehat w_{i',\cdot}\|_2/ \|\widehat w_{i',\cdot}\|_1))(\sum_{j \in \NC[i']} \widehat w_{i',j})+ \frac{2}{m}\sum_{i' = 1}^n \rho_{i', i} (\sum_{j \in \NC[i']} \widehat w_{i',j})  \\
    & +  \frac{2}{m}\sum_{j=1}^m\frac{C_{\delta}\sigma^2}{\K[j]}+\sum_{i' = 1}^n\frac{2\drown{i'}{i}}{m}\brackets{\sum_{j\in \NC}\frac{1}{\K}} -\sum_{i' = 1}^n\frac{2\drown{i'}{i}}{m}(\sum_{j\in \NC}\widehat w_{i', j}) \\
    \stackrel{\ref{assump:dist_concentration}}{\leq}  & \frac{2}{m} \sum_{i' = 1}^n (16M^2 \sqrt{\log(2/\delta)} (\|\widehat w_{i',\cdot}\|_2/ \|\widehat w_{i',\cdot}\|_1))(\sum_{j \in \NC[i']} \widehat w_{i',j})+\sum_{i' = 1}^n  \frac{2 \widehat \rho_{i', i}}{m} (\sum_{j \in \NC[i']} \widehat w_{i',j}) + \sum_{i' = 1}^n \frac{2 \zeta_{i', i}(m , \delta)}{m} (\sum_{j \in \NC[i']} \widehat w_{i',j}) \\
    & +  \frac{2}{m}\sum_{j=1}^m\frac{C_{\delta}\sigma^2}{\K[j]}+\sum_{i' = 1}^n\frac{2\drown{i'}{i}}{m}\brackets{\sum_{j\in \NC}\frac{1}{\K}} -\sum_{i' = 1}^n\frac{2\drown{i'}{i}}{m}(\sum_{j\in \NC}\widehat w_{i', j}) \\
    = & \frac{2}{m}\sum_{j=1}^m\frac{C_{\delta}\sigma^2}{\K[j]}+\sum_{i' = 1}^n\frac{2\drown{i'}{i}}{m}\brackets{\sum_{j\in \NC}\frac{1}{\K}} \\
    & + \frac{2}{m}\sum_{i' = 1}^n \left(16M^2 \sqrt{\log(2/\delta)} (\|\widehat w_{i',\cdot}\|_2/ \|\widehat w_{i',\cdot}\|_1) + \zeta_{i', i}(m , \delta) \right) (\sum_{j\in \NC}\widehat w_{i', j}) \\
    = & \frac{2}{m}\sum_{j=1}^m\frac{C_{\delta}\sigma^2}{\K[j]}+ \frac{2}{m} \sum_{j = 1}^m \frac{\sum_{i' \in \NR}\drown{i'}{i}}{\K} \\
    & + \frac{2}{m} \sum_{j=1}^m \sum_{i' \in \NR} \widehat w_{i',j} \left(16M^2 \sqrt{\log(2/\delta)} (\|\widehat w_{i',\cdot}\|_2/ \|\widehat w_{i',\cdot}\|_1) + \zeta_{i', i}(m , \delta) \right) . 
\end{align}

This completes the proof of the theorem.

\newpage

\newpage
\end{document}